\documentclass{article}
\usepackage{iclr2027_conference,times}

\usepackage{amsmath,amsfonts,bm}

\def\eqref#1{equation~\ref{#1}}

\def\plaineqref#1{\ref{#1}}

\def\1{\bm{1}}

\DeclareMathAlphabet{\mathsfit}{\encodingdefault}{\sfdefault}{m}{sl}
\SetMathAlphabet{\mathsfit}{bold}{\encodingdefault}{\sfdefault}{bx}{n}

\usepackage{hyperref}
\usepackage{amssymb}
\usepackage{url}
\usepackage{booktabs}
\usepackage{multirow}
\usepackage{colortbl}
\usepackage{tabularx}
\usepackage{graphicx}
\usepackage{algorithm}
\usepackage{algpseudocode}

\definecolor{denserow}{RGB}{244,244,244}
\definecolor{parkrow}{RGB}{236,236,252}

\title{PARK: Accurate Block Retrieval for Sparse Attention in Video Diffusion Transformers}

\author{
\textbf{Yun Dai}$^{1}$\quad
\textbf{Jiarui Wen}$^{1}$\quad
\textbf{Huiping Zhuang}$^{1}$\quad
\textbf{Cen Chen}$^{1}$\quad
\textbf{Ziqian Zeng}$^{1}$\thanks{Corresponding author.} \\
$^{1}$South China University of Technology \\
\texttt{zqzeng@scut.edu.cn}
}

\iclrfinalcopy
\begin{document}

\setcounter{footnote}{1}
\maketitle
\lhead{\footnotesize PARK: Accurate Block Retrieval for Sparse Attention in Video Diffusion Transformers}

\begin{abstract}
Diffusion Transformers (DiTs) have become a dominant architecture for video generation, but their efficiency is limited by the quadratic complexity of full attention.
Sparse attention reduces this cost by retrieving important blocks and computing attention only within them, but inaccurate retrieval can either degrade generation quality or yield unnecessary computation.
We identify two retrieval mismatches in methods that retrieve blocks using the averaged representations of query and key blocks: 
(i) query-side aggregation mismatch, where averaging queries before Softmax fails to preserve their individual attention preferences, 
and (ii) key-side clustering metric mismatch, where standard Euclidean clustering in the original key space can group keys with dissimilar QK scores under the current query, so their average representation may not accurately represent how the current query scores individual keys.
These mismatches can lead to inaccurate block retrieval.
To address these mismatches, we propose PARK, a training-free sparse attention method for accurate block retrieval.
PARK retains every original query, independently normalizes its attention over key blocks, and then averages these distributions within each query block. 
It also uses information from the current queries to transform keys before clustering, so that keys receiving similar QK scores are grouped together.
A fused GPU kernel further reduces the overhead of block retrieval. 
Experiments on HunyuanVideo and Wan demonstrate that PARK improves block retrieval accuracy and preserves generation quality while accelerating inference, achieving the best quality-efficiency trade-off among the compared sparse attention methods.

\end{abstract}

\section{Introduction}

Diffusion Transformers (DiTs) (\cite{dit2023}) have emerged as a dominant paradigm for modern video generation, demonstrating strong scalability and enabling high-fidelity, temporally coherent synthesis (\cite{cogvideox,hunyuanvideo,wan}). 
However, generating high-resolution and long videos involves a large number of spatio-temporal tokens.
Since 3D full attention scales quadratically with token count, it becomes a major computational bottleneck in video generation (\cite{svg,turbodiffusion,attention2017}).

Fortunately, attention maps in video DiTs are inherently sparse, with only a small fraction of token interactions contributing significantly to the output (\cite{vorta,vsa}). 
Sparse attention exploits this property by computing attention only over these critical token interactions, substantially reducing computation with little degradation in generation quality. 
In practice, query and key tokens are partitioned into blocks by grouping consecutive tokens, and attention is computed only for a selected subset of blocks.
These blocks are retrieved by estimating their importance and selecting those with the highest scores (\cite{block_sparse_attn,mminference}).
This process, referred to as block retrieval, produces a \textit{sparse mask} that specifies which blocks are computed.

Existing training-free sparse attention methods broadly fall into static and dynamic methods.
Static methods (\cite{compact,svg,radial,sta}) derive sparse masks from offline attention profiling or predefined rules and apply them to individual layers or attention heads during inference. 
Dynamic methods (\cite{adaspa,spargeattention,xattention}) perform block retrieval at runtime using lightweight block importance estimation, adapting the sparse masks to the current attention pattern. 
Additionally, recent token-reordering methods (\cite{svg2,paroattention,adacluster,svoo}) cluster semantically similar tokens and permute them into contiguous blocks.
This helps retain more critical interactions with fewer blocks, improving the efficiency-quality trade-off.

Even with token reordering, the effectiveness of sparse attention depends on how accurately it retrieves important blocks.
Missing important blocks can degrade generation quality, while selecting unimportant blocks wastes computation.
To compensate for inaccurate retrieval, more blocks may need to be retained to preserve generation quality, limiting the achievable speedup.
Accurate block retrieval is therefore essential to improving the efficiency-quality trade-off of sparse attention.

\begin{figure}[t]
    \centering
    \includegraphics[width=\linewidth]{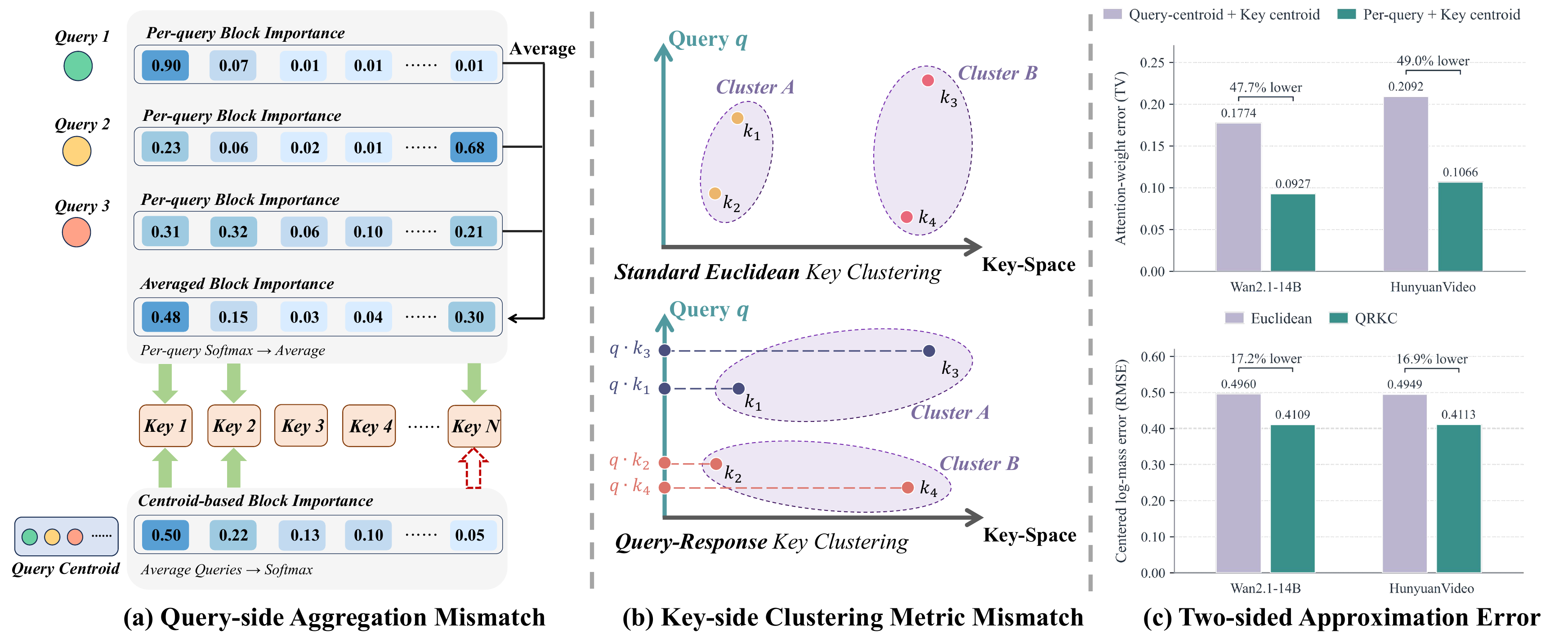}
    \vspace{-20pt}
    \caption{
        Two-sided retrieval mismatches and approximation errors in
        centroid-based block retrieval.
        (a) Averaging queries into a single centroid before
        Softmax normalization ignores their individual key preferences,
        leading to inaccurate block retrieval.
        (b) Standard Euclidean clustering considers only distances between keys in feature space
        and does not ensure that keys within the same cluster receive similar scores from the current query, 
        leading to inaccurate key-centroid estimates.
        (c) Retaining the original queries instead of using query centroids
        (top) and clustering keys by QK-score similarity (bottom) both reduce approximation errors.
        More experimental details are provided in Appendix~\ref{app:mismatch_experiments}.
    }
    \label{fig:retrieval_mismatches}
    \vspace{-7pt}
\end{figure}

To keep block retrieval efficient, many existing methods estimate block importance from the average-pooled representations (or \textit{centroids}) of query and key blocks, 
Our theoretical analysis of the resulting estimation errors reveals two retrieval mismatches in this common centroid-based design:
\begingroup
\setlength{\leftmargini}{2pc}
\begin{itemize}
    \item \textbf{M1: Query-side aggregation mismatch.}
    Each query independently yields an attention distribution over all key blocks, and these query-specific preferences jointly determine which blocks should be selected for the sparse mask.
    As analyzed in Sec.~\ref{sec:query_aggregation_mismatch}, averaging the original queries into a single query centroid before Softmax normalization ignores these individual preferences and can misestimate the attention assigned to each key block, leading to an inaccurate sparse mask.
    Figure~\ref{fig:retrieval_mismatches}(c) shows that retaining the original queries instead of using query centroids reduces attention-weight approximation error by 47.7--49.0\%.
    \item \textbf{M2: Key-side clustering metric mismatch.}
    Existing reordering methods cluster key tokens using standard Euclidean clustering in the original key space and use the key block centroids for importance estimation.
    As analyzed in Sec.~\ref{sec:key_clustering_mismatch}, centroid-based estimation is more accurate when keys in the same block receive similar dot-product scores from the current query.
    However, this clustering metric compares only the keys themselves and does not account for how the current query scores them.
    Consequently, standard Euclidean clustering may group keys that are scored differently by the current query, making key centroids less accurate for block importance estimation.
    Figure~\ref{fig:retrieval_mismatches}(c) shows that clustering keys by QK-score similarity reduces centered log-mass error by 16.9--17.2\% compared with standard Euclidean clustering.
\end{itemize}
\endgroup

To address these mismatches, we propose \underline{\textbf{PARK}}, a training-free sparse attention method for accurate block retrieval via \underline{\textbf{P}}er-query \underline{\textbf{A}}ttention-Mass Aggregation (PAMA) and Query-\underline{\textbf{R}}esponse \underline{\textbf{K}}ey Clustering (QRKC).
First, PAMA retains original queries, computes a separate Softmax distribution over the key blocks for each query, and aggregates these distributions by averaging them within each query block to estimate block importance for sparse mask selection.
This preserves query-specific attention preferences and eliminates the aggregation mismatch introduced by query centroids.
We fuse per-query normalization and averaging into a custom kernel for efficient execution.
Second, QRKC uses a linear transformation derived from the current queries to map the keys into a query-conditioned space for clustering, where Euclidean distance reflects how similarly those queries score them.
By grouping keys that are scored similarly by the current queries, QRKC reduces the approximation error introduced by key centroids.
Together, PAMA and QRKC improve block retrieval with negligible overhead. Our contributions are summarized as follows:

\begin{itemize}
    \item We identify two retrieval mismatches in existing centroid-based block retrieval methods: query-side aggregation mismatch and key-side clustering metric mismatch, and analyze how they introduce errors in block-importance estimation.
    
    \item We propose PARK, a novel training-free sparse attention method for fast video generation. It addresses the two identified mismatches through Per-query Attention-Mass Aggregation and Query-Response Key Clustering.

    \item Extensive experiments on HunyuanVideo and Wan demonstrate that PARK consistently improves block retrieval accuracy and preserves generation quality with high end-to-end speedup, achieving the best quality-efficiency trade-off among the compared training-free sparse attention methods.
\end{itemize}

\section{Related Work}

\paragraph{Training-based Sparse Attention.}Training-based sparse attention methods fine-tune video diffusion models with task-specific sparse attention patterns.
VSA (\cite{vsa}) employs a coarse-to-fine routing mechanism to select important attention blocks, while VMoBA (\cite{vmoba}) cycles through 1D, 2D, and 3D block partitions across layers and retrieves blocks using a global similarity threshold.
Other methods, such as DSV (\cite{dsv}), BSA (\cite{bsa}), and MoGA (\cite{moga}), train lightweight predictors, routers, or cluster assignments to estimate sparse attention patterns before computation.
However, they generally require additional training or fine-tuning and may introduce auxiliary prediction modules.

\paragraph{Training-free Sparse Attention.}Training-free methods directly exploit attention sparsity in pretrained video DiTs without additional optimization.
Static methods (\cite{svg,sta,compact,radial}) derive sparse masks from offline profiling or predefined spatio-temporal structures.
Dynamic methods (\cite{minference,spargeattention,xattention,adaspa}) instead construct sparse masks at runtime by estimating the importance of attention blocks.
Recent methods further improve sparse attention through token reordering.
PAROAttention (\cite{paroattention}) performs pattern-aware token reordering for sparse and quantized attention.
SVG2 (\cite{svg2}) and AdaCluster (\cite{adacluster}) apply K-means clustering to query and key tokens and reorder semantically similar tokens into contiguous blocks, while SVOO (\cite{svoo}) incorporates query--key interactions through bidirectional co-clustering.

\section{Preliminary}

\subsection{Attention and Sparse Attention}

For a single attention head, the input tokens are projected into query, key, and value matrices $\mathbf{Q},\mathbf{K},\mathbf{V}\in\mathbb{R}^{L\times d}$, where $L$ is the sequence length and $d$ is the head dimension.
Full attention (\cite{attention2017,flashattention}) is computed as
\begin{equation}
    \mathbf{Z}=\frac{\mathbf{Q}\mathbf{K}^{\top}}{\sqrt{d}},\qquad
    \mathbf{A}=\operatorname{Softmax}(\mathbf{Z}),\qquad
    \mathbf{O}=\mathbf{A}\mathbf{V},
    \label{eq:full_attention}
\end{equation}
where the Softmax is applied along the key dimension independently for each query.
Computing the full attention matrix incurs quadratic complexity with respect to the sequence length.

Given a sparse mask $\mathbf{M}\in\{0,1\}^{L\times L}$, sparse attention is computed as
\begin{equation}
    \mathbf{O}^{\mathrm{sp}}
    =(\mathbf{A}\odot\mathbf{M})\mathbf{V},
    \label{eq:sparse_attention}
\end{equation}
where $\odot$ denotes element-wise masking, and $M_{ij}=0$ indicates that the interaction from query $i$ to key $j$ is skipped.

In practice, query and key tokens are partitioned into query blocks $\{\mathcal{Q}_u\}_{u=1}^{N_Q}$ and key blocks $\{\mathcal{K}_v\}_{v=1}^{N_K}$, respectively.
Each query--key block consists of all attention interactions between the queries in $\mathcal{Q}_u$ and the keys in $\mathcal{K}_v$, corresponding to a submatrix of $\mathbf{A}$.
The sparse mask selects or skips each query--key block as a whole.
The block sizes are denoted by $n_u^Q=|\mathcal{Q}_u|$ and $n_v^K=|\mathcal{K}_v|$, with $\sum_{u=1}^{N_Q}n_u^Q=\sum_{v=1}^{N_K}n_v^K=L$.

\subsection{Block Importance and Attention Recall}

For a query block $\mathcal{Q}_u$ and a key block $\mathcal{K}_v$, we define their exact block importance ${W}_{uv}$ as the average attention weight that the queries in $\mathcal{Q}_u$ assign to $\mathcal{K}_v$:
\begin{equation}
    W_{uv}
    =\frac{1}{n_u^Q}
    \sum_{i\in\mathcal{Q}_u}
    \sum_{j\in\mathcal{K}_v} A_{ij}.
    \label{eq:block_importance}
\end{equation}
Here, $A_{ij}$ denotes the dense post-Softmax attention weight from query token $i$ to key token $j$.
For every query block, the block importance is normalized over all key blocks, i.e., $\sum_{v=1}^{N_K}W_{uv}=1$.
Under Top-$p$ retrieval, key blocks are retrieved for each query block until their cumulative estimated importance reaches the threshold $p$.
In practice, key blocks are retrieved in descending order of $\widehat{W}_{uv}$, which estimates the exact block importance $W_{uv}$.
The retained query--key block pairs form the block-level sparse mask $\mathbf{M}$.

The block retrieval accuracy of sparse attention is measured using attention recall, defined as
\begin{equation}
    \operatorname{Recall}(\mathbf{A},\mathbf{M})
    =
    \frac{
        \sum_{i=1}^{L}\sum_{j=1}^{L}A_{ij}M_{ij}
    }{
        \sum_{i=1}^{L}\sum_{j=1}^{L}A_{ij}
    }.
    \label{eq:attention_recall}
\end{equation}
Attention recall (\cite{recall}) measures the proportion of full-attention weights covered by the block-level sparse mask.
A higher recall indicates more accurate block retrieval.

\section{Motivation}
\label{sec:motivation}

Centroid-based block retrieval compresses both query and key blocks before estimating their importance.
Although this design is efficient, the two compressions introduce fundamentally different approximations.
We analyze them relative to the exact block importance in Eq.~(\plaineqref{eq:block_importance}) and show how they lead to the two retrieval mismatches.

\subsection{Query-Side Aggregation Mismatch}
\label{sec:query_aggregation_mismatch}
Averaging queries into a centroid before Softmax can lead to inaccurate block-importance estimates.
To analyze this error, we retain all original keys.
The exact block importance $W_{uv}$ in Eq.~(\plaineqref{eq:block_importance}) is computed by first normalizing attention for each query and then averaging the attention weights of queries in the same block.
Query-centroid estimation instead replaces these queries with their centroid before Softmax normalization.
A Taylor expansion expresses the approximation error as
\begin{equation}
    \Delta_{uv}^{Q}
    =
    W_{uv}-\widehat{W}_{uv}^{\mathrm{centroid}}
    =\frac{1}{2}\left\langle
    \mathbf{H}_{\phi_v}(\overline{\mathbf{q}}_u),
    \boldsymbol{\Sigma}_{Q,u}
    \right\rangle_F
    +\mathcal{R}_{uv}^{Q}.
    \label{eq:query_pooling_gap_summary}
\end{equation}
Here, $\widehat{W}_{uv}^{\mathrm{centroid}}=\phi_v(\overline{\mathbf{q}}_u)$ is the query-centroid estimate, where $\phi_v(\mathbf{q})$ is the exact attention weight that query $\mathbf{q}$ assigns to key block $\mathcal{K}_v$, and $\overline{\mathbf{q}}_u$ is the query-block centroid.
$\boldsymbol{\Sigma}_{Q,u}$ is the within-block query covariance, and $\mathbf{H}_{\phi_v}(\overline{\mathbf{q}}_u)$ is the Hessian of $\phi_v$, capturing how $\phi_v$ curves as the query varies near its centroid.
$\langle\cdot,\cdot\rangle_F$ denotes the Frobenius inner product, and $\mathcal{R}_{uv}^{Q}$ is the higher-order remainder.
The full derivation is provided in Appendix~\ref{app:query_pooling_derivation}.

Accurate query-centroid estimation requires a small $|\Delta_{uv}^{Q}|$, whose leading error is determined by $\langle \mathbf{H}_{\phi_v}(\overline{\mathbf{q}}_u),\boldsymbol{\Sigma}_{Q,u}\rangle_F$.
A sufficient condition for a small leading error is that the product of the norm of $\mathbf{H}_{\phi_v}(\overline{\mathbf{q}}_u)$ and the total within-block query variance is small.
Standard Euclidean query clustering reduces the total within-block query variance, captured by $\boldsymbol{\Sigma}_{Q,u}$, but does not generally eliminate it.
Since query clustering does not account for the Hessian $\mathbf{H}_{\phi_v}(\overline{\mathbf{q}}_u)$, it does not by itself ensure that this product is sufficiently small.
Empirically, Table~\ref{tab:query_representation_errors} shows that query-centroid estimation with full keys incurs even higher approximation error than retaining original queries with key centroids.

We further evaluate whether retaining the original queries remains beneficial when keys are represented by centroids.
Figure~\ref{fig:retrieval_mismatches}(c) shows that retaining original queries reduces approximation error when both estimates use the same key centroids.
Table~\ref{tab:query_key_compression} further shows that retaining full queries with key centroids achieves the highest recall among the compressed representations across both ordering schemes and models.
These empirical results support preserving query-specific attention preferences in practical block retrieval and motivate Per-Query Attention-Mass Aggregation (§\ref{sec:pama}).

\begin{table}[t]
    \centering
    \caption{Top-$p$ block-retrieval recall ($p=0.9$) under different query/key representations. Mean, P05, and Worst denote the mean, fifth percentile, and minimum attention recall across all attention heads and layers, respectively. All values are percentages. Higher is better.}
    \label{tab:query_key_compression}
    \small
    \setlength{\tabcolsep}{4.0pt}
    \renewcommand{\arraystretch}{1.08}
    \begin{tabularx}{\linewidth}{@{}lll*{6}{>{\centering\arraybackslash}X}@{}}
        \toprule
        \multirow{2}{*}{Ordering} & \multirow{2}{*}{Query} & \multirow{2}{*}{Key}
        & \multicolumn{3}{c}{HunyuanVideo (\cite{hunyuanvideo})} & \multicolumn{3}{c}{Wan 2.1-14B (\cite{wan})} \\
        \cmidrule(lr){4-6}\cmidrule(l){7-9}
        & & & Mean & P05 & Worst & Mean & P05 & Worst \\
        \midrule
        \rowcolor{denserow}
        \cellcolor{white} & Full & Full & 91.29 & 90.02 & 90.01 & 91.71 & 90.02 & 90.01 \\
        & Centroid & Centroid & 88.06 & 81.53 & 64.21 & 90.43 & 87.11 & 73.89 \\
        & Centroid & Full & 89.10 & 84.44 & 68.40 & 91.00 & 88.64 & 79.06 \\
        \rowcolor{parkrow}
        \cellcolor{white}\multirow{-4}{*}{Clustered}
        & Full & Centroid
        & \textbf{90.70} & \textbf{87.62} & \textbf{81.64}
        & \textbf{91.56} & \textbf{89.36} & \textbf{83.61} \\
        \midrule
        \rowcolor{denserow}
        \cellcolor{white} & Full & Full & 90.64 & 90.02 & 90.01 & 90.89 & 90.03 & 90.02 \\
        & Centroid & Centroid & 85.59 & 72.40 & 22.94 & 90.01 & 83.96 & 69.11 \\
        & Centroid & Full & 86.18 & 56.17 & 15.02 & 89.24 & 74.92 & 39.94 \\
        \rowcolor{parkrow}
        \cellcolor{white}\multirow{-4}{*}{Original}
        & Full & Centroid
        & \textbf{87.80} & \textbf{76.35} & \textbf{24.30}
        & \textbf{91.32} & \textbf{87.02} & \textbf{73.55} \\
        \bottomrule
    \end{tabularx}
\end{table}

\subsection{Key-Side Clustering Metric Mismatch}
\label{sec:key_clustering_mismatch}

With the original queries retained, the exact attention weight $\phi_v(\mathbf{q}_i)$ is obtained by applying Softmax to the block log-masses:
$\phi_v(\mathbf{q}_i)=\exp(S_{iv}^{\star})/
\sum_{r=1}^{N_K}\exp(S_{ir}^{\star})$, where $S_{iv}^{\star}$ is the exact log-mass of key block $\mathcal{K}_v$ for query $\mathbf{q}_i$.
The exact block log-mass and its key-centroid estimate are
\begin{equation}
    S_{iv}^{\star}
    =\log\sum_{j\in\mathcal{K}_v}
    \exp\!\left(\frac{\mathbf{q}_i^{\top}\mathbf{k}_j}{\sqrt{d}}\right),
    \qquad
    \widehat{S}_{iv}
    =\log n_v^K+\frac{\mathbf{q}_i^{\top}\mathbf{c}_v}{\sqrt{d}}.
    \label{eq:key_logmass_estimates}
\end{equation}
Here, $\mathbf{c}_v=(1/n_v^K)\sum_{j\in\mathcal{K}_v}\mathbf{k}_j$ is the key-block centroid, and $n_v^K$ is the number of keys in $\mathcal{K}_v$.
The term $\log n_v^K$ accounts for key-block size.
An expansion expresses the key-centroid approximation error as
\begin{equation}
    \Delta_{iv}^{K}
    =S_{iv}^{\star}-\widehat{S}_{iv}
    =\frac{1}{2d}
    \mathbf{q}_i^{\top}\boldsymbol{\Sigma}_{K,v}\mathbf{q}_i
    +\mathcal{R}_{iv}^{K}.
    \label{eq:key_logmass_summary}
\end{equation}
Here, $\boldsymbol{\Sigma}_{K,v}$ is the within-block key covariance, $\epsilon_{iv}^{K}=\frac{1}{2d}\mathbf{q}_i^{\top}\boldsymbol{\Sigma}_{K,v}\mathbf{q}_i$ is the leading term of $\Delta_{iv}^{K}$, and $\mathcal{R}_{iv}^{K}$ is the higher-order remainder.
The full derivation is provided in Appendix~\ref{app:key_metric_derivation}.

Accurate key-centroid log-mass estimation requires a small $|\Delta_{iv}^{K}|$.
The leading term $\epsilon_{iv}^{K}$ is proportional to the within-block key variance after projection onto the current query, which is the variance of the unscaled QK scores within that block.
Standard euclidean clustering aims to reduce within-block key variation, captured by $\boldsymbol{\Sigma}_{K,v}$, but does not account for the current query $\mathbf{q}_i$ and therefore does not directly minimize $\epsilon_{iv}^{K}$.
Additional empirical evidence for this metric mismatch is provided in Appendix~\ref{app:key_score_errors}.
Therefore, key clustering should account for QK-score similarity under the current query, motivating Query-Response Key Clustering (§\ref{sec:qrkc}).

\begin{figure}[t]
    \centering
    \includegraphics[width=\linewidth]{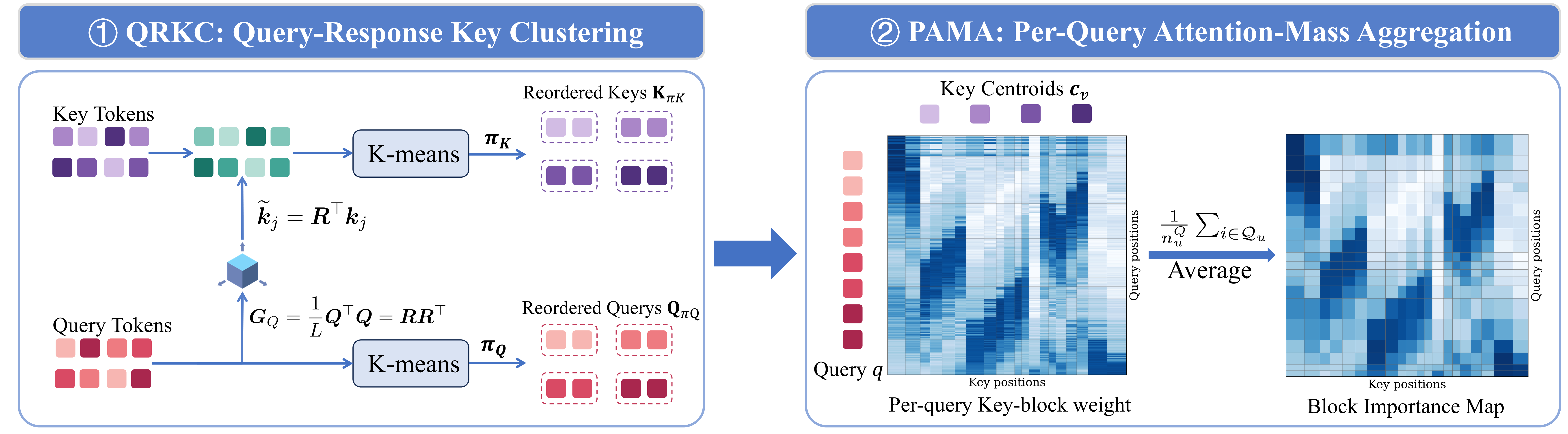}
    \vspace{-20pt}
    \caption{
        Overview of PARK.
        QRKC transforms keys using a metric derived from the current queries, so that keys receiving similar QK scores are grouped together, while queries are clustered using standard Euclidean K-means.
        PAMA retains every original query, computes its Softmax-normalized weights over the key centroids, and averages them within each query block to produce the block-importance map.
    }
    \label{fig:overview}
    \vspace{-5pt}
\end{figure}

\section{Methodology}
In this section, we introduce PARK, a training-free sparse attention framework designed for accurate block retrieval via Per-Query Attention-Mass Aggregation (PAMA) and Query-Response Key Clustering (QRKC).
PAMA preserves all original queries, computes their normalized attention over key blocks separately, and then averages these attention weights within each query block for block-importance estimation (§\ref{sec:pama}).
QRKC uses information from the current queries to transform the keys before clustering, so that keys receiving similar QK scores are grouped together (§\ref{sec:qrkc}).
These two modules improve block retrieval on both the query and key sides.
An overview of our proposed PARK framework is shown in Figure~\ref{fig:overview}.

\subsection{PAMA: Per-Query Attention-Mass Aggregation}
\label{sec:pama}
As shown in Eq.~(\plaineqref{eq:block_importance}), exact block importance is computed by normalizing attention independently for every query and then averaging the attention weights within each query block.
Following this order, PAMA retains all original queries during block-importance estimation.
For each key block $\mathcal{K}_v$, we compute its centroid $\mathbf{c}_v$ in the original key space and approximate its log-mass $\widehat{S}_{iv}$ for query $\mathbf{q}_i$ as
\begin{equation}
    \mathbf{c}_v
    =
    \frac{1}{n_v^K}
    \sum_{j\in\mathcal{K}_v}\mathbf{k}_j,
    \qquad
    \widehat{S}_{iv}
    =
    \frac{\mathbf{q}_i^{\top}\mathbf{c}_v}{\sqrt{d}}
    +
    \log n_v^K.
    \label{eq:pama_block_score}
\end{equation}
Here, $\mathcal{K}_v$ denotes a key block containing $n_v^K$ keys.
The block-size term $\log n_v^K$ accounts for the different numbers of keys represented by the block centroids.

PAMA then normalizes the estimated scores over all key blocks for each query and averages the block weights within query block $\mathcal{Q}_u$:
\begin{equation}
    \widehat{P}_{iv}
    =
    \frac{\exp(\widehat{S}_{iv})}
    {\sum_{r=1}^{N_K}\exp(\widehat{S}_{ir})},
    \qquad
    \widehat{W}_{uv}^{\mathrm{PAMA}}
    =
    \frac{1}{n_u^Q}
    \sum_{i\in\mathcal{Q}_u}
    \widehat{P}_{iv}.
    \label{eq:pama_block_importance}
\end{equation}
Here, $N_K$ is the number of key blocks, and $\mathcal{Q}_u$ is a query block containing $n_u^Q$ queries.
$\widehat{P}_{iv}$ is the estimated attention weight that query $\mathbf{q}_i$ assigns to key block $\mathcal{K}_v$, and $\widehat{W}_{uv}^{\mathrm{PAMA}}$ is the importance of the query--key block.

\paragraph{Efficient Kernel Implementation.}
For each attention head, PAMA requires $\mathcal{O}(LN_Kd)$ operations for QK-centroid scores and $\mathcal{O}(LN_K)$ for row-wise Softmax and averaging within query blocks, far below full attention's $\mathcal{O}(L^2d)$ cost since $N_K\ll L$.
A straightforward implementation stores $L\times N_K$ logits and probabilities in HBM, incurring $\mathcal{O}(LN_K)$ intermediate storage and additional memory access.
To reduce this memory overhead, we fuse the QK-centroid matrix multiplication, row-wise Softmax, and averaging within query blocks into a single GPU kernel.
The intermediate logits and probabilities remain in on-chip registers, and only the final $N_Q\times N_K$ block-importance matrix is written to HBM.
The pseudocode and overhead analysis are provided in Appendix~\ref{app:fused_pama_kernel}.

\subsection{QRKC: Query-Response Key Clustering}
\label{sec:qrkc}

QRKC aims to reduce the leading key-centroid approximation error $\epsilon_{iv}^{K}$ identified in Sec.~\ref{sec:key_clustering_mismatch}.
To construct a key-clustering objective that accounts for all current queries, we average the squared QK-score error caused by replacing keys with their centroids over these queries:
\begin{equation}
    \mathcal{J}_{\mathrm{QRKC}}
    =\frac{1}{L}\sum_{i=1}^{L}\sum_{v=1}^{N_K}
    \sum_{j\in\mathcal{K}_v}
    \left[\mathbf{q}_i^\top(\mathbf{k}_j-\mathbf{c}_v)\right]^2
    =\sum_{v=1}^{N_K}\sum_{j\in\mathcal{K}_v}
    \|\mathbf{k}_j-\mathbf{c}_v\|_{\mathbf{G}_Q}^{2}.
    \label{eq:qrkc_objective}
\end{equation}
Here, $\mathbf{G}_Q=(1/L)\mathbf{Q}^{\top}\mathbf{Q}\in\mathbb{R}^{d\times d}$ is the uncentered second-moment matrix of the current queries, and $\|\mathbf{x}\|_{\mathbf{G}_Q}^{2}=\mathbf{x}^{\top}\mathbf{G}_Q\mathbf{x}$.
The relation $\mathcal{J}_{\mathrm{QRKC}} =2d\sum_v n_v^K(1/L)\sum_i\epsilon_{iv}^{K}$ connects this clustering objective $\mathcal{J}_{\mathrm{QRKC}}$ directly to the leading key-centroid approximation error $\epsilon_{iv}^{K}$.

To minimize $\mathcal{J}_{\mathrm{QRKC}}$ efficiently, we reuse a standard Euclidean K-means implementation.
However, standard K-means uses the squared Euclidean distance
$\|\mathbf{k}_j-\mathbf{c}_v\|_2^2$, whereas our objective in Eq.~(\plaineqref{eq:qrkc_objective}) uses $\|\mathbf{k}_j-\mathbf{c}_v\|_{\mathbf{G}_Q}^2=(\mathbf{k}_j-\mathbf{c}_v)^\top\mathbf{G}_Q(\mathbf{k}_j-\mathbf{c}_v)$.
We therefore transform the keys into a space where the squared Euclidean distance between each key and its centroid satisfies the following equality:
\begin{equation}
    \|\widetilde{\mathbf{k}}_j-\widetilde{\mathbf{c}}_v\|_2^2
    =(\mathbf{k}_j-\mathbf{c}_v)^{\top}\mathbf{G}_Q
    (\mathbf{k}_j-\mathbf{c}_v).
    \label{eq:qrkc_transformed_distance}
\end{equation}
Here, $\widetilde{\mathbf{k}}_j$ and $\widetilde{\mathbf{c}}_v$ denote a transformed key and its cluster centroid, respectively.
The sum of these squared distances over all keys equals $\mathcal{J}_{\mathrm{QRKC}}$ in Eq.~(\plaineqref{eq:qrkc_objective}).
Thus, standard Euclidean K-means in the transformed space directly minimizes this objective $\mathcal{J}_{\mathrm{QRKC}}$.

Directly mapping each key to $\mathbf{Q}\mathbf{k}_j/\sqrt{L}$ also realizes this distance in Eq.~(\plaineqref{eq:qrkc_transformed_distance}), but increases the feature dimension from $d$ to $L$.
To construct a transformation that avoids this increase in feature dimension while satisfying Eq.~(\plaineqref{eq:qrkc_transformed_distance}), we apply Cholesky decomposition and transform each key as
\begin{equation}
    \mathbf{G}_Q=\mathbf{R}\mathbf{R}^{\top},
    \qquad
    \widetilde{\mathbf{k}}_j=\mathbf{R}^{\top}\mathbf{k}_j
    \in\mathbb{R}^{d}.
    \label{eq:qrkc_key_transform}
\end{equation}
Here, $\mathbf{R}$ is a Cholesky factor when $\mathbf{G}_Q$ is positive definite.
Since the transformation is linear, $\widetilde{\mathbf{c}}_v=\mathbf{R}^{\top}\mathbf{c}_v$ is the centroid of the transformed keys in block $v$.
This transformation satisfies the distance equality in Eq.~(\plaineqref{eq:qrkc_transformed_distance}).
For numerical stability, we use a regularized metric, which yields the corresponding regularized clustering objective.
The derivation and stabilization details are provided in Appendix~\ref{app:key_metric_derivation}, and the QRKC pseudocode is provided in Appendix~\ref{app:qrkc_pseudocode}.

\paragraph{Clustering and Reuse Strategy.}
We apply standard Euclidean K-means to cluster and reorder the queries, while using QRKC to cluster and reorder the keys.
Prior studies~(\cite{dfsattn,svoo,adaspa}) have shown that sparse masks and clustering results change little across adjacent diffusion steps.
To reduce online clustering and block-retrieval overhead, we reuse the sparse masks and clustering results across diffusion steps and recompute them every \(R\) steps.

\paragraph{Difference from SVG2 and SVOO.}
SVG2 (\cite{svg2}) independently clusters queries and keys using Euclidean K-means, while SVOO (\cite{svoo}) couples the two partitions through iterative bidirectional co-clustering.
Both methods then estimate block importance using query and key centroids.
In contrast, PAMA retains every original query, normalizes its attention over key blocks separately, and then averages these weights within each query block to estimate block importance.
QRKC derives a query-dependent metric from all current queries and transforms the keys before clustering, so that Euclidean K-means minimizes the average squared QK-score error introduced by key centroids.
PARK therefore adopts an asymmetric retrieval design that preserves query-specific preferences while clustering keys using a theoretically derived metric.

\section{Experiments}
\subsection{Experimental Setup}
\label{sec:experimental_setup}
\textbf{Models.} We evaluate PARK on four open-source DiT models, including Wan2.1-T2V-1.3B, Wan2.1-T2V-14B, Wan2.2-T2V-14B (\cite{wan}), and HunyuanVideo-T2V-13B (\cite{hunyuanvideo}).
All models generate videos at 720p resolution.
In latent space, the Wan2.1 models process 21 frames, whereas HunyuanVideo processes 32 frames, with 3,600 tokens per frame in all cases.   

\textbf{Metrics.} We use PSNR (\cite{psnr}), SSIM (\cite{ssim}), and LPIPS (\cite{lpips}) to measure the similarity between videos generated with sparse and dense attention.
For video quality evaluation, we report Subject Consistency, Background Consistency, Image Quality, and Aesthetic Quality from VBench (\cite{vbench}).
To assess efficiency, we report the end-to-end latency and the overall speedup.

\textbf{Datasets.} For all the text-to-video generation experiments, we use the original prompts from VBench (\cite{vbench}) for both HunyuanVideo and Wan.

\textbf{Baselines.} We compare PARK with state-of-the-art training-free sparse attention methods, including SpargeAttention (\cite{spargeattention}), XAttention (\cite{xattention}), SVG (\cite{svg}), SVG2 (\cite{svg2}), and SVOO (\cite{svoo}).
All baselines are evaluated under the same hardware configuration and input settings as PARK.

\textbf{Implementations.} We set the number of query blocks to \(N_Q=128\) for all models, and the number of key blocks to \(N_K=1{,}024\) for HunyuanVideo and \(N_K=512\) for Wan.
All models generate videos with 50 denoising steps.
We implement the fused PAMA kernel with Triton (\cite{triton}) and use a FlashInfer (\cite{flashinfer}) kernel with dynamic block sizes for sparse attention computation.
All models use dense attention during the first 20\% of the diffusion steps and a fixed Top-\(p\) threshold of \(p=0.9\).
We recompute sparse masks and clustering results every \(R=10\) diffusion steps.
All experiments are conducted on a single NVIDIA A800 GPU.  

\begin{table}[t]
    \centering
    \caption{Quality and efficiency benchmark results for PARK and the baselines.}
    \label{tab:main_quality_efficiency}
    \scriptsize
    \setlength{\tabcolsep}{2.6pt}
    \renewcommand{\arraystretch}{1.05}
    \resizebox{\linewidth}{!}{
    \begin{tabular}{@{}llccccccccc@{}}
        \toprule
        \multirow{2}{*}{Model} & \multirow{2}{*}{Method}
        & \multicolumn{7}{c}{Quality} & \multicolumn{2}{c}{Efficiency} \\
        \cmidrule(lr){3-9}\cmidrule(l){10-11}
        & & PSNR$\uparrow$ & SSIM$\uparrow$ & LPIPS$\downarrow$
        & SubCons.$\uparrow$ & BackCons.$\uparrow$ & ImageQual.$\uparrow$ & AesQual.$\uparrow$
        & Latency$\downarrow$ & Speedup$\uparrow$ \\
        \midrule
        \rowcolor{denserow}
        \cellcolor{white}\multirow{7}{*}{\shortstack[l]{Wan2.2-14B}}
        & Dense       & -- & -- & -- & 97.13\% & 97.22\% & 73.78\% & 65.56\% & 3792s & 1.00$\times$ \\
        & SpargeAttn  & 18.32 & 0.714 & 0.295 & 97.10\% & \underline{97.33\%} & 73.73\% & 65.55\% & \underline{2614s} & \underline{1.45$\times$} \\
        & XAttention  & 17.31 & 0.640 & 0.353 & 96.89\% & 97.10\% & \textbf{73.87\%} & \textbf{66.07\%} & 2819s & 1.34$\times$ \\
        & SVG         & 17.33 & 0.663 & 0.342 & 96.99\% & 96.96\% & 73.73\% & \underline{65.85\%} & 2668s & 1.42$\times$ \\
        & SVG2        & 18.77 & 0.730 & 0.288 & 96.88\% & 97.11\% & 73.71\% & 65.20\% & 2672s & 1.42$\times$ \\
        & SVOO        & \underline{19.08} & \underline{0.753} & \underline{0.267} & \textbf{97.15\% }& \textbf{97.34\%} & \underline{73.76\%} & 65.31\% & 2722s & 1.39$\times$ \\
        \rowcolor{parkrow}
        \cellcolor{white} & \textbf{PARK} & \textbf{19.19} & \textbf{0.754} & \textbf{0.265} & \underline{97.11\%} & 97.23\% & 73.73\% & 65.29\% & \textbf{2452s} & \textbf{1.55$\times$} \\
        \midrule
        \rowcolor{denserow}
        \cellcolor{white}\multirow{7}{*}{\shortstack[l]{Wan2.1-1.3B}}
        & Dense       & -- & -- & -- & 97.36\% & 96.99\% & 70.59\% & 61.23\% & 745s & 1.00$\times$ \\
        & SpargeAttn  & 25.15 & 0.857 & 0.231 & \underline{97.28\%} & \underline{96.95\%} & \underline{71.44\%} & \textbf{61.58\%} & 478s & 1.57$\times$ \\
        & XAttention  & 23.43 & 0.786 & 0.296 & 96.68\% & 96.36\% & 68.71\% & 60.82\% & 495s & 1.51$\times$ \\
        & SVG         & 22.39 & 0.754 & 0.336 & 96.73\% & 96.47\% & \textbf{72.16\%} & 60.99\% & 506s & 1.47$\times$ \\
        & SVG2        & 25.98 & 0.867 & 0.230 & 96.45\% & 96.03\% & 67.78\% & 58.90\% & \underline{435s} & \underline{1.71$\times$} \\
        & SVOO        & \underline{27.36} & \underline{0.901} & \underline{0.187} & \textbf{97.32\%} & \textbf{97.07\%} & 70.94\% & \underline{61.25\%} & 472s & 1.58$\times$ \\
        \rowcolor{parkrow}
        \cellcolor{white} & \textbf{PARK} & \textbf{27.52} & \textbf{0.908} & \textbf{0.177} & 97.22\% & 96.87\% & 70.48\% & 61.02\% & \textbf{423s} & \textbf{1.76$\times$} \\
        \midrule
        \rowcolor{denserow}
        \cellcolor{white}\multirow{7}{*}{\shortstack[l]{Wan2.1-14B}}
        & Dense       & -- & -- & -- & 97.95\% & 97.83\% & 72.46\% & 61.83\% & 3595s & 1.00$\times$ \\
        & SpargeAttn  & 20.55 & 0.757 & 0.295 & 94.88\% & 95.49\% & \textbf{72.80\%} & 61.73\% & 2732s & 1.32$\times$ \\
        & XAttention  & 21.67 & 0.794 & 0.266 & 97.82\% & 97.73\% & 71.94\% & \underline{62.13\%} & 2710s & 1.33$\times$ \\
        & SVG         & 20.87 & 0.785 & 0.277 & 97.74\% & \textbf{97.90\%} & 71.99\% & \textbf{63.18\%} & 2747s & 1.31$\times$ \\
        & SVG2        & 22.98 & 0.848 & 0.219 & 97.82\% & 97.51\% & 72.24\% & 61.61\% & \underline{2490s} & \underline{1.44$\times$} \\
        & SVOO        & \underline{23.66} & \underline{0.871} & \underline{0.188} & \underline{97.89\%} & \underline{97.76\%} & \underline{72.42\%} & 61.63\% & 2754s & 1.31$\times$ \\
        \rowcolor{parkrow}
        \cellcolor{white} & \textbf{PARK} & \textbf{24.01} & \textbf{0.875} & \textbf{0.186} & \textbf{97.91\%} & 97.74\% & 72.28\% & 62.03\% & \textbf{2354s} & \textbf{1.53$\times$} \\
        \midrule
        \rowcolor{denserow}
        \cellcolor{white}\multirow{7}{*}{\shortstack[l]{HunyuanVideo}}
        & Dense       & -- & -- & -- & 96.95\% & 96.98\% & 69.69\% & 55.58\% & 3058s & 1.00$\times$ \\
        & SpargeAttn  & 29.32 & 0.923 & 0.187 & 96.83\% & \underline{96.44\%} & \textbf{69.90\%} & \textbf{55.44\%} & 2064s & 1.48$\times$ \\
        & XAttention  & 29.08 & 0.909 & 0.198 & 96.80\% & 96.28\% & \underline{69.44\%} & 55.27\% & 2243s & 1.36$\times$ \\
        & SVG         & 27.88 & 0.898 & 0.221 & 96.71\% & 96.38\% & 68.96\% & \underline{55.32\%} & 2066s & 1.48$\times$ \\
        & SVG2        & 29.27 & 0.914 & 0.213 & 96.60\% & 96.05\% & 66.50\% & 54.05\% & 1978s & \underline{1.55$\times$} \\
        & SVOO        & \underline{30.31} & \underline{0.931} & \underline{0.172} & \textbf{96.88\%} & 96.39\% & 69.19\% & 55.05\% & \underline{1975s} & \underline{1.55$\times$} \\
        \rowcolor{parkrow}
        \cellcolor{white} & \textbf{PARK} & \textbf{30.69} & \textbf{0.936} & \textbf{0.166} & \underline{96.87\%} & \textbf{96.50\%} & 69.41\% & 55.07\% & \textbf{1837s} & \textbf{1.66$\times$} \\
        \bottomrule
    \end{tabular}
    }
\end{table}

\subsection{Quality and Efficiency Evaluation}
\begin{figure}[t]
    \centering
    \includegraphics[width=\linewidth]{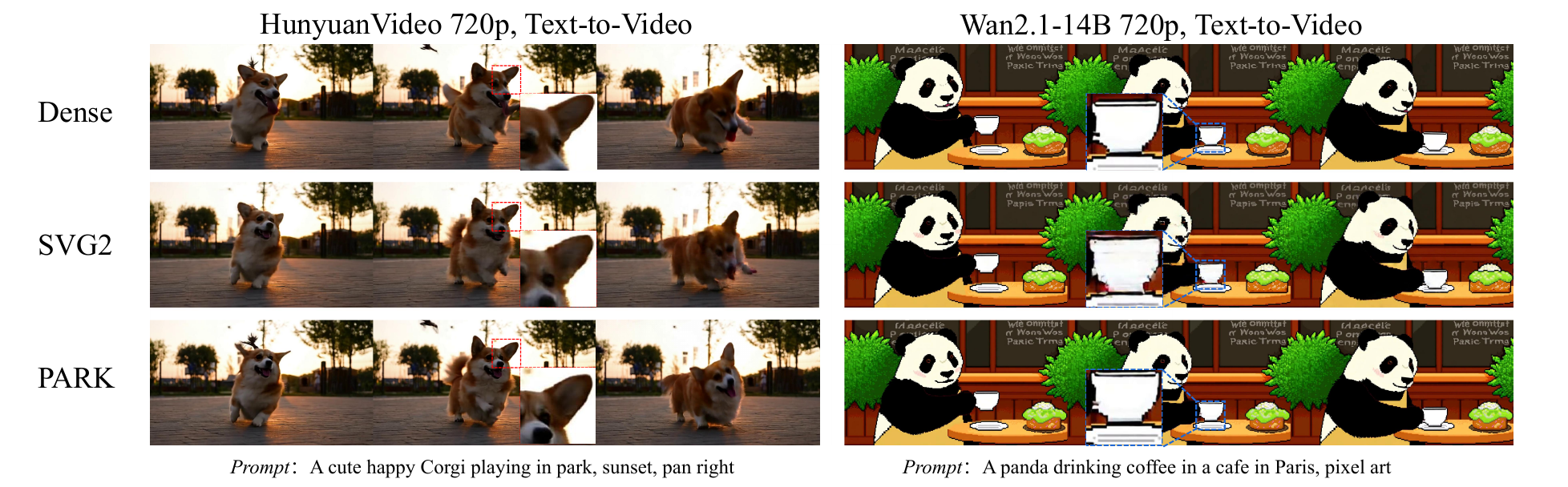}
    \vspace{-18pt}
    \caption{
        Examples of videos generated by PARK and SVG2 on HunyuanVideo and Wan2.1-14B.
    }
    \label{fig:generation_example}
    \vspace{-7pt}
\end{figure}

As shown in Table~\ref{tab:main_quality_efficiency}, PARK consistently achieves the best similarity to dense attention across all four models, attaining the highest PSNR and SSIM and the lowest LPIPS among the compared sparse attention methods.
It also remains close to dense attention on the VBench metrics, showing that PARK preserves high generation quality.
Compared with SVG2, PARK improves Image Quality by $2.91$ and $2.70$ percentage points on HunyuanVideo and Wan2.1-1.3B, respectively, with smaller improvements on Wan2.1-14B and Wan2.2-14B.
As shown in Figure~\ref{fig:generation_example}, PARK produces videos that more closely match dense attention than SVG2, with better-preserved fine-grained visual details.

While preserving this level of generation quality, PARK delivers $1.55\times$, $1.76\times$, $1.53\times$, and $1.66\times$ end-to-end speedups over dense attention on Wan2.2-14B, Wan2.1-1.3B, Wan2.1-14B, and HunyuanVideo, respectively, achieving the highest speedup among the compared methods on all four models.
Overall, PARK achieves the best quality-efficiency trade-off among the compared sparse-attention methods.

\begin{table}[H]
    \vspace{-10pt}
    \centering
    \caption{Ablation study of PAMA and QRKC on HunyuanVideo and Wan2.1-14B.}
    \label{tab:module_ablation}
    \small
    \setlength{\tabcolsep}{5.0pt}
    \renewcommand{\arraystretch}{1.08}
    \begin{tabularx}{\linewidth}{@{}ll*{4}{>{\centering\arraybackslash}X}@{}}
        \toprule
        Model
        & Variant
        & PSNR $\uparrow$
        & SSIM $\uparrow$
        & LPIPS $\downarrow$
        & Latency $\downarrow$ \\
        \midrule

        \rowcolor{blue!5}
        \cellcolor{white} & PARK       & \textbf{29.85} & \textbf{0.931} & \textbf{0.168} & 1827s \\
        \cellcolor{white} & w/o PAMA   & 28.40 & 0.911 & 0.205 & 1824s \\
        \cellcolor{white}\multirow{-3}{*}{HunyuanVideo}
        & w/o QRKC   & 29.54 & 0.928 & 0.173 & 1826s \\

        \midrule

        \rowcolor{blue!5}
        \cellcolor{white} & PARK       & \textbf{23.90} & \textbf{0.871} & \textbf{0.189} & 2359s \\
        \cellcolor{white} & w/o PAMA   & 23.48 & 0.864 & 0.199 & 2353s \\
        \cellcolor{white}\multirow{-3}{*}{Wan2.1-14B}
        & w/o QRKC   & 23.80 & 0.869 & 0.190 & 2354s \\

        \bottomrule
    \end{tabularx}
\end{table}

\subsection{Ablation Study}
We evaluate the contributions of PAMA and QRKC on a fixed subset of VBench prompts by removing each module individually, while keeping the model configurations and inference settings identical to those used in the main evaluation.
Removing PAMA replaces per-query estimation with query-centroid estimation, while removing QRKC replaces query-response key clustering with standard Euclidean K-means.

As shown in Table~\ref{tab:module_ablation}, removing PAMA consistently degrades PSNR, SSIM, and LPIPS, especially on HunyuanVideo, confirming the importance of preserving query-specific attention preferences during block retrieval.
Removing QRKC also reduces similarity to dense attention on both models, demonstrating the benefit of grouping keys according to their QK-score similarity under the current queries.
Meanwhile, the latency differences among the three variants remain within \(0.3\%\), indicating that PAMA and QRKC improve generation quality from the query and key sides, respectively, with negligible overhead.

\section{Conclusion}
In this paper, we proposed PARK, a training-free sparse attention method designed for fast diffusion-based video generation.
We identified query-side aggregation mismatch and key-side clustering metric mismatch in centroid-based block retrieval.
To address them, PAMA preserves query-specific attention preferences, while QRKC clusters keys using a query-dependent metric.
Experiments across four models show that PARK preserves high generation quality while achieving $1.53\times$--$1.76\times$ end-to-end speedups over dense attention.

\bibliography{iclr2027_conference}
\bibliographystyle{iclr2027_conference}

\newpage
\appendix

\section{Direct Evaluation of Retrieval Mismatches}
\label{app:mismatch_experiments}

\begin{figure}[H]
    \centering
    \includegraphics[width=\linewidth]
    {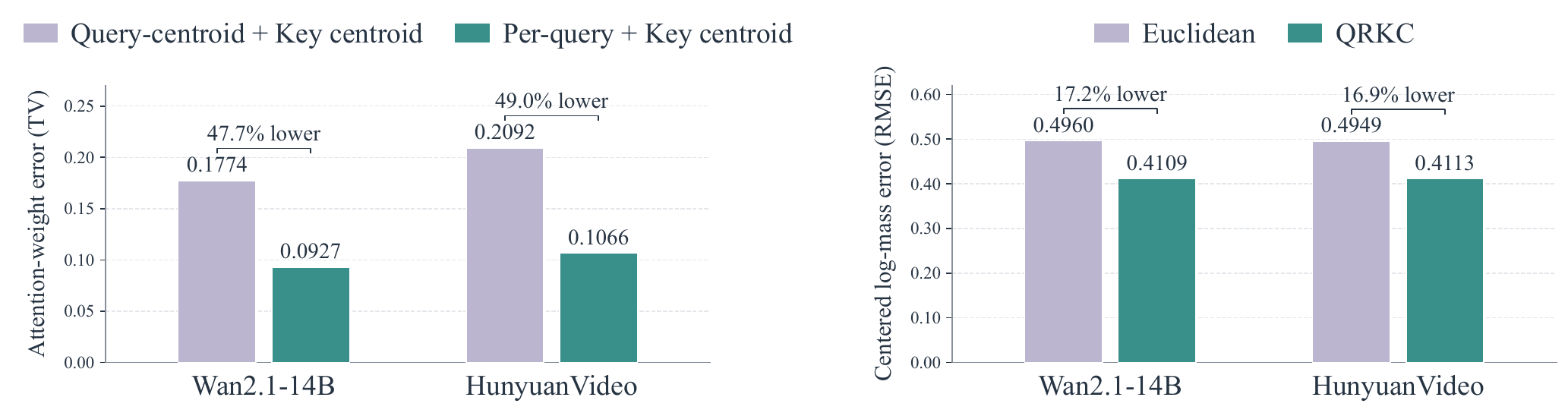}
    \vspace{-10pt}
    \caption{
        Direct evaluation of the two retrieval mismatches.
        Left: retaining original queries reduces attention-weight
        approximation error compared with query-centroid estimation.
        Right: QRKC reduces centered log-mass error compared with
        standard Euclidean clustering.
        Results correspond to Figure~\ref{fig:retrieval_mismatches}(c).
        Lower is better.
    }
    \label{fig:mismatch_errors_appendix}
\end{figure}

We evaluate both mismatches on Wan2.1-14B and HunyuanVideo. 
We measure approximation errors against exact full-query, full-key references, with the results summarized in Figure~\ref{fig:mismatch_errors_appendix}.

\paragraph{M1: Attention-weight approximation error.}
We fix the Euclidean query and key partitions and compare per-query and query-centroid estimation using the same key centroids. Unlike the full-key analysis in Section~\ref{sec:query_aggregation_mismatch}, this experiment includes key-centroid approximation in both estimates.
The per-query estimation normalizes attention separately for each query before averaging the block weights, whereas the query-centroid estimation averages queries before normalization.
Both estimates are evaluated against the exact block importance $W_{uv}$ in Eq.~(\plaineqref{eq:block_importance}), computed from the full queries and keys.
For each layer--head pair, we measure
\begin{equation}
    E_{\mathrm{TV}}^{m}
    =\sum_{u=1}^{N_Q}\frac{n_u^Q}{L}
    \left(\frac{1}{2}\sum_{v=1}^{N_K}
    \left|\widehat{W}_{uv}^{m}-W_{uv}\right|\right),
    \qquad m\in\{\mathrm{per},\mathrm{centroid}\}.
    \label{eq:appendix_mismatch_tv}
\end{equation}
Here, $E_{\mathrm{TV}}^{m}$ is the mean attention-weight TV error for one layer--head pair, with each query block weighted by its number of queries.
The superscripts $m=\mathrm{per}$ and $m=\mathrm{centroid}$ denote per-query and query-centroid estimation, respectively, and $\widehat{W}_{uv}^{m}$ is the corresponding block-importance estimate.
For each query block $u$, $\frac{1}{2}\sum_{v=1}^{N_K}|\widehat{W}_{uv}^{m}-W_{uv}|$ is the total variation (TV) distance between the estimated and exact block-weight distributions.
We average the resulting errors equally across layer--head pairs. 
This metric compares each estimate with the exact reference, rather than measuring the difference between the two estimates.

As shown in Figure~\ref{fig:mismatch_errors_appendix}, retaining the original queries reduces mean TV from $0.1774$ to $0.0927$ on Wan2.1-14B and from $0.2092$ to $0.1066$ on HunyuanVideo. 
These correspond to relative reductions of $47.7\%$ and $49.0\%$, respectively. 
These results show that using query centroids introduces substantial additional approximation error even when the key representation is unchanged.

\paragraph{Query-centroid error with full keys.}
To directly evaluate the full-key analysis in Section~\ref{sec:query_aggregation_mismatch}, we additionally retain all original keys and replace only the queries with their block centroids.
For this configuration, the estimated block importance is $\widehat{W}_{uv}=\phi_v(\overline{\mathbf{q}}_u)$, where $\phi_v$ uses all original keys as defined in Eq.~(\plaineqref{eq:app_exact_block_weight}).
We use the same Euclidean query and key partitions, exact reference $W_{uv}$, and TV weighting and averaging as above.
Table~\ref{tab:query_representation_errors} includes the two key-centroid configurations from Figure~\ref{fig:mismatch_errors_appendix} for comparison.

\begin{table}[!htbp]
    \centering
    \caption{Mean attention-weight TV under different query and key representations, using the same Euclidean partitions. Full-query, full-key estimation is the exact reference and has zero error by definition. The two key-centroid rows reproduce the M1 results in Figure~\ref{fig:mismatch_errors_appendix}. Lower is better. Bold indicates the lowest error among the compressed representations.}
    \label{tab:query_representation_errors}
    \small
    \setlength{\tabcolsep}{8pt}
    \begin{tabular}{llcc}
        \toprule
        Query & Key & Wan2.1-14B & HunyuanVideo \\
        \midrule
        \rowcolor{denserow}
        Full & Full & 0.0000 & 0.0000 \\
        Centroid & Centroid & 0.1774 & 0.2092 \\
        Centroid & Full & 0.1335 & 0.1667 \\
        \rowcolor{parkrow}
        Full & Centroid & \textbf{0.0927} & \textbf{0.1066} \\
        \bottomrule
    \end{tabular}
\end{table}

As shown in Table~\ref{tab:query_representation_errors}, query-centroid estimation incurs mean TV errors of $0.1335$ and $0.1667$ even without key compression, directly measuring the approximation error analyzed in Section~\ref{sec:query_aggregation_mismatch}.
Retaining the original queries while compressing keys yields lower mean errors than compressing queries while retaining full keys, with relative reductions of $30.6\%$ and $36.0\%$ on Wan2.1-14B and HunyuanVideo, respectively.

\vspace{30pt}
\paragraph{M2: Centered log-mass approximation error.}
We retain all original queries and compare Euclidean key clustering with QRKC. For each partition, we compute the exact log-mass $S_{iv}^{\star}$ using the log-sum-exp expression in Eq.~(\plaineqref{eq:key_logmass_summary}) and its key-centroid estimate $\widehat{S}_{iv}$ using Eq.~(\plaineqref{eq:pama_block_score}). Their difference gives the approximation error:
\begin{equation}
    \Delta_{iv}^{K}=S_{iv}^{\star}-\widehat{S}_{iv}.
    \label{eq:appendix_mismatch_logmass_error}
\end{equation}
For each query, we subtract the mean error across key blocks, since a common additive offset does not affect Softmax. We then measure the centered error using
\begin{equation}
    \delta_{iv}
    =\Delta_{iv}^{K}-\frac{1}{|\mathcal V|}\sum_{r\in\mathcal V}\left(S_{ir}^{\star}-\widehat{S}_{ir}\right),
    \qquad
    E_{\mathrm{LM}}
    =\frac{1}{L}\sum_{i=1}^{L}
    \sqrt{\frac{1}{|\mathcal V|}\sum_{v\in\mathcal V}\delta_{iv}^{2}}.
    \label{eq:appendix_mismatch_centered_rmse}
\end{equation}
Here, $E_{\mathrm{LM}}$ is the mean centered log-mass RMSE over all queries for one layer--head pair.
$\mathcal V$ is the set of nonempty key blocks, and $r$ is the index of these blocks in the sum. 
Thus, $\delta_{iv}$ subtracts the mean log-mass error over all nonempty key blocks for query $i$ from its error on block $v$. 
We compute $E_{\mathrm{LM}}$ across key blocks separately for each query, average over the $L$ queries, and then average equally across layer--head pairs. 
Each method uses the exact reference under its own key partition.

As shown in Figure~\ref{fig:mismatch_errors_appendix}, QRKC reduces mean centered log-mass RMSE from $0.4960$ to $0.4109$ on Wan2.1-14B and from $0.4949$ to $0.4113$ on HunyuanVideo, corresponding to relative reductions of $17.2\%$ and $16.9\%$, respectively. 
These results show that clustering keys by QK-score similarity improves log-mass approximation compared with Euclidean clustering.

\section{Key-Space and QK-Score Approximation Errors}
\label{app:key_score_errors}
Using the same configuration as Appendix~\ref{app:mismatch_experiments}, we compare the errors caused by replacing each key with its cluster centroid in the original key space and in the scaled QK scores:
\begin{equation}
D_{\mathrm{key}}=\frac{1}{L_K}\sum_{j=1}^{L_K}
\|\mathbf{k}_j-\mathbf{c}_{v(j)}\|_2^2,
\qquad
D_{\mathrm{QK}}=\frac{1}{L_Q L_K}\sum_{i=1}^{L_Q}\sum_{j=1}^{L_K}
\left[\frac{\mathbf{q}_i^\top(\mathbf{k}_j-\mathbf{c}_{v(j)})}{\sqrt d}\right]^2.
\label{eq:key_score_errors}
\end{equation}
Here, $v(j)$ is the cluster containing key $j$, and $L_Q$ and $L_K$ are the numbers of queries and keys. 
Raw-key MSE $D_{\mathrm{key}}$ averages squared Euclidean distances over keys, without dividing by the feature dimension. 
$D_{\mathrm{QK}}$ averages squared scaled-score errors over all query--key pairs.
Equivalently, it averages the within-block score variance over queries, weighting key blocks by their sizes. 
It therefore corresponds to twice the similarly weighted leading error term $\epsilon_{iv}^{K}$, not the full log-mass error. 
Both metrics are averaged equally across layer--head pairs.

\begin{table}[H]
    \centering
    \vspace{-7pt}
    \caption{Key-space and QK-score errors. Changes are relative to Euclidean clustering.}
    \label{tab:key_score_errors}
    \small
    \setlength{\tabcolsep}{6pt}
    \begin{tabular}{@{}llrrr@{}}
        \toprule
        Model & Metric & Euclidean & QRKC & Change \\
        \midrule
        Wan2.1-14B & Raw-key MSE & 66.88 & 72.59 & $+8.5\%$ \\
        \rowcolor{denserow}
        \cellcolor{white} & QK-score squared error & 11.49 & \textbf{7.28} & $\mathbf{-36.6\%}$ \\
        \midrule
        HunyuanVideo & Raw-key MSE & 79.51 & 91.31 & $+14.8\%$ \\
        \rowcolor{denserow}
        \cellcolor{white} & QK-score squared error & 2.92 & \textbf{2.26} & $\mathbf{-22.8\%}$ \\
        \bottomrule
    \end{tabular}
\end{table}
\vspace{-7pt}
Table~\ref{tab:key_score_errors} shows that QRKC increases raw-key MSE while reducing QK-score error on both models. 
Thus, smaller Euclidean key-centroid distances do not necessarily yield more accurate QK scores, supporting the key-side clustering metric mismatch analyzed in Sec.~\ref{sec:key_clustering_mismatch}.

\section{Effect of Query-Response Key Clustering}
\begin{figure}[H]
    \centering
    \includegraphics[width=\linewidth]{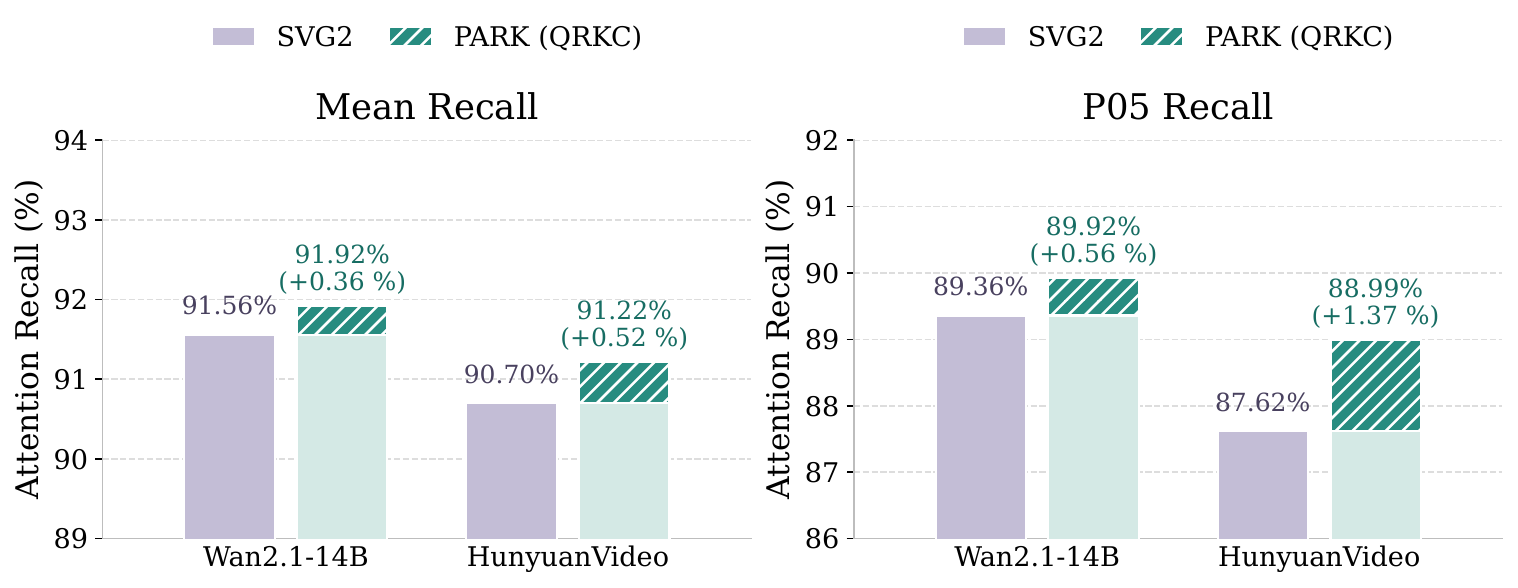}
    \vspace{-15pt}
    \caption{
        Effect of QRKC on mean and P05 attention recall.
    }
    \label{fig:qrkc_effect}
\end{figure}

We evaluate QRKC by replacing Euclidean key clustering while keeping PAMA and the other settings unchanged.
As shown in Figure~\ref{fig:qrkc_effect}, QRKC improves both mean and P05 attention recall on Wan2.1-14B and HunyuanVideo.
These results support clustering keys according to their QK-score similarity under the current queries for more accurate block retrieval.

\newpage

\section{Detailed Derivations for Centroid-Based Block Retrieval}
\label{app:derivations}

This appendix provides the complete expansions underlying the analyses in Sections~\ref{sec:query_aggregation_mismatch} and~\ref{sec:key_clustering_mismatch}.
We retain all original keys when analyzing query-centroid approximation, and retain all original queries when analyzing key-centroid approximation. 

The query and key partitions are held unchanged within each comparison. All sums over blocks below concern nonempty blocks. 
The quantities $\Delta_{uv}^{Q}$, $\epsilon_{iv}^{K}$, and $\Delta_{iv}^{K}$ have the same meanings as in the main text: the exact block importance minus its query-centroid estimate, the leading key-side error term, and the full key-side log-mass error, respectively.

\subsection{Derivation of the Query-Side Aggregation Mismatch}
\label{app:query_pooling_derivation}

Replacing queries with their centroid can introduce error even when all keys are retained exactly.
Following Section~\ref{sec:query_aggregation_mismatch}, we analyze this error without any key-side compression.
The exact attention weight assigned to a key block is
\begin{equation}
    \phi_v(\mathbf{q})
    =\sum_{j\in\mathcal{K}_v}p_j(\mathbf{q}),
    \qquad
    p_j(\mathbf{q})
    =\frac{\exp(\mathbf{q}^{\top}\mathbf{k}_j/\sqrt d)}
    {\sum_{\ell=1}^{L}\exp(\mathbf{q}^{\top}\mathbf{k}_{\ell}/\sqrt d)}.
    \label{eq:app_exact_block_weight}
\end{equation}
Here, $p_j(\mathbf{q})$ is the exact token-level attention weight for key $\mathbf{k}_j$, and $\phi_v(\mathbf{q})$ sums these weights over key block $\mathcal{K}_v$.
The denominator includes all $L$ original keys, and $\ell$ indexes individual keys.
The query and key vectors belong to $\mathbb{R}^{d}$.
Thus, averaging the exact block weights over the original queries gives the ground truth, whereas evaluating them at the query centroid gives an approximation:
\begin{equation}
    W_{uv}
    =\frac{1}{n_u^Q}\sum_{i\in\mathcal{Q}_u}\phi_v(\mathbf{q}_i),
    \qquad
    \widehat{W}_{uv}^{\mathrm{centroid}}
    =\phi_v(\overline{\mathbf{q}}_u).
    \label{eq:app_query_estimators}
\end{equation}
Here, $W_{uv}$ is the exact block importance in Eq.~(\plaineqref{eq:block_importance}), and $\widehat{W}_{uv}^{\mathrm{centroid}}$ is its query-centroid estimate.
The set $\mathcal{Q}_u$ contains $n_u^Q$ queries, and $\overline{\mathbf{q}}_u$ is their centroid.
Both expressions use every original key. The only approximation in this comparison is replacing the queries with their centroid before Softmax normalization. Per-query computation itself is exact.

The query-block statistics used in the expansion are
\begin{equation}
    \overline{\mathbf{q}}_u
    =\frac{1}{n_u^Q}\sum_{i\in\mathcal{Q}_u}\mathbf{q}_i,
    \qquad
    \boldsymbol{\epsilon}_i
    =\mathbf{q}_i-\overline{\mathbf{q}}_u.
    \label{eq:app_query_residual}
\end{equation}
Here, $\overline{\mathbf{q}}_u$ is the arithmetic mean of the queries and $\boldsymbol{\epsilon}_i$ is the vector deviation of query $i$ from that mean. This vector is distinct from the scalar key-side error term $\epsilon_{iv}^{K}$.
The deviations have zero mean. Indeed,
\begin{equation}
    \frac{1}{n_u^Q}\sum_{i\in\mathcal{Q}_u}\boldsymbol{\epsilon}_i
    =\frac{1}{n_u^Q}\sum_{i\in\mathcal{Q}_u}\mathbf{q}_i
    -\overline{\mathbf{q}}_u
    =\overline{\mathbf{q}}_u-\overline{\mathbf{q}}_u
    =\mathbf{0}.
    \label{eq:app_query_zero_mean}
\end{equation}
Their second moment is
\begin{equation}
    \boldsymbol{\Sigma}_{Q,u}
    =\frac{1}{n_u^Q}\sum_{i\in\mathcal{Q}_u}
    \boldsymbol{\epsilon}_i\boldsymbol{\epsilon}_i^{\top}
    \in\mathbb{R}^{d\times d}.
    \label{eq:app_query_covariance}
\end{equation}
Here, $\boldsymbol{\Sigma}_{Q,u}$ is the within-block query covariance, normalized by $n_u^Q$ rather than the sample-covariance denominator $n_u^Q-1$.

\paragraph{Taylor expansion of the exact block importance.}
For finite key vectors, $\phi_v$ in Eq.~(\plaineqref{eq:app_exact_block_weight}) is smooth because its denominator is strictly positive.
We expand it along the segments connecting the query centroid to each query in $\mathcal{Q}_u$.
The derivatives at the centroid are
\begin{equation}
    \mathbf{g}_v
    =\nabla_{\mathbf{q}}\phi_v(\overline{\mathbf{q}}_u),
    \qquad
    \mathbf{H}_v
    =\nabla_{\mathbf{q}}^2\phi_v(\overline{\mathbf{q}}_u).
    \label{eq:app_query_derivatives}
\end{equation}
Here, $\mathbf{g}_v$ is the gradient and $\mathbf{H}_v=\mathbf{H}_{\phi_v}(\overline{\mathbf{q}}_u)$ is exactly the Hessian appearing in Eq.~(\plaineqref{eq:query_pooling_gap_summary}). 
Their dependence on the query block $u$ is omitted from the notation for simplicity.
A second-order Taylor expansion of each $\phi_v(\mathbf{q}_i)$ about the common point $\overline{\mathbf{q}}_u$ gives
\begin{equation}
    \phi_v(\mathbf{q}_i)
    =\phi_v(\overline{\mathbf{q}}_u)
    +\mathbf{g}_v^{\top}\boldsymbol{\epsilon}_i
    +\frac{1}{2}\boldsymbol{\epsilon}_i^{\top}
    \mathbf{H}_v\boldsymbol{\epsilon}_i
    +\mathcal{R}_{iv}^{Q,(3)}.
    \label{eq:app_query_taylor}
\end{equation}
Here, $\mathcal{R}_{iv}^{Q,(3)}$ is the higher-order remainder.
Taylor's theorem gives
\begin{equation}
    \left|\mathcal{R}_{iv}^{Q,(3)}\right|
    \leq \frac{M_v}{6}\|\boldsymbol{\epsilon}_i\|_2^3.
    \label{eq:app_query_pointwise_remainder_bound}
\end{equation}
Here, $M_v$ is a uniform upper bound on the operator norm of the third derivative $\nabla^3\phi_v$ along all segments connecting $\overline{\mathbf{q}}_u$ to the queries in $\mathcal{Q}_u$. 
The same bound $M_v$ applies to all queries in $\mathcal{Q}_u$, so it also bounds the remainder after averaging within the block.

Substituting Eq.~(\plaineqref{eq:app_query_taylor}) into the definition of the exact block importance $W_{uv}$ yields
\begin{equation}
    \begin{aligned}
        W_{uv}
        ={}&\phi_v(\overline{\mathbf{q}}_u)
        +\mathbf{g}_v^{\top}
        \left(\frac{1}{n_u^Q}\sum_{i\in\mathcal{Q}_u}\boldsymbol{\epsilon}_i\right)\\
        &+\frac{1}{2n_u^Q}\sum_{i\in\mathcal{Q}_u}
        \boldsymbol{\epsilon}_i^{\top}\mathbf{H}_v\boldsymbol{\epsilon}_i
        +\frac{1}{n_u^Q}\sum_{i\in\mathcal{Q}_u}\mathcal{R}_{iv}^{Q,(3)}.
    \end{aligned}
    \label{eq:app_query_row_expansion}
\end{equation}
Equation~(\plaineqref{eq:app_query_zero_mean}) proves explicitly that the gradient term vanishes:
\begin{equation}
    \mathbf{g}_v^{\top}
    \left(\frac{1}{n_u^Q}\sum_{i\in\mathcal{Q}_u}\boldsymbol{\epsilon}_i\right)
    =\mathbf{g}_v^{\top}\mathbf{0}=0.
    \label{eq:app_query_gradient_cancellation}
\end{equation}
For the quadratic term, the identity
$\boldsymbol{\epsilon}_i^{\top}\mathbf{H}_v\boldsymbol{\epsilon}_i
=\operatorname{tr}(\mathbf{H}_v\boldsymbol{\epsilon}_i\boldsymbol{\epsilon}_i^{\top})$
gives
\begin{equation}
    \frac{1}{n_u^Q}\sum_{i\in\mathcal{Q}_u}
    \boldsymbol{\epsilon}_i^{\top}\mathbf{H}_v\boldsymbol{\epsilon}_i
    =\operatorname{tr}(\mathbf{H}_v\boldsymbol{\Sigma}_{Q,u})
    =\langle\mathbf{H}_v,\boldsymbol{\Sigma}_{Q,u}\rangle_F.
    \label{eq:app_query_trace_identity}
\end{equation}
Here, $\operatorname{tr}(\cdot)$ is the matrix trace and
$\langle\mathbf{A},\mathbf{B}\rangle_F=\operatorname{tr}(\mathbf{A}^{\top}\mathbf{B})$
is the Frobenius inner product.
Because a Hessian is symmetric, $\operatorname{tr}(\mathbf{H}_v\boldsymbol{\Sigma}_{Q,u})
=\langle\mathbf{H}_v,\boldsymbol{\Sigma}_{Q,u}\rangle_F$.

Averaging the pointwise remainders gives
\begin{equation}
    \mathcal{R}_{uv}^{Q}
    =\frac{1}{n_u^Q}\sum_{i\in\mathcal{Q}_u}
    \mathcal{R}_{iv}^{Q,(3)}.
    \label{eq:app_query_aggregate_remainder}
\end{equation}
Here, $\mathcal{R}_{uv}^{Q}$ is the higher-order remainder used in the main text.
Combining Eqs.~(\plaineqref{eq:app_query_row_expansion})--(\plaineqref{eq:app_query_trace_identity}), the exact block importance expands as
\begin{equation}
    W_{uv}
    =\phi_v(\overline{\mathbf{q}}_u)
    +\frac{1}{2}\langle\mathbf{H}_v,\boldsymbol{\Sigma}_{Q,u}\rangle_F
    +\mathcal{R}_{uv}^{Q}.
    \label{eq:app_query_row_final}
\end{equation}
In contrast, the query-centroid estimator evaluates $\phi_v$ exactly at the expansion point and is simply
\begin{equation}
    \widehat{W}_{uv}^{\mathrm{centroid}}
    =\phi_v(\overline{\mathbf{q}}_u).
    \label{eq:app_query_pool_final}
\end{equation}
Subtracting Eq.~(\plaineqref{eq:app_query_pool_final}) from Eq.~(\plaineqref{eq:app_query_row_final}) now gives
\begin{equation}
    \Delta_{uv}^{Q}
    =W_{uv}
    -\widehat{W}_{uv}^{\mathrm{centroid}}
    =\frac{1}{2}\langle\mathbf{H}_v,\boldsymbol{\Sigma}_{Q,u}\rangle_F
    +\mathcal{R}_{uv}^{Q}.
    \label{eq:app_query_pooling_gap}
\end{equation}
Using the uniform bound in Eq.~(\plaineqref{eq:app_query_pointwise_remainder_bound}) and the triangle inequality, the averaged remainder satisfies
\begin{equation}
    |\mathcal{R}_{uv}^{Q}|
    \leq\frac{M_v}{6n_u^Q}
    \sum_{i\in\mathcal{Q}_u}\|\boldsymbol{\epsilon}_i\|_2^3.
    \label{eq:app_query_remainder_bound}
\end{equation}
Thus, the first-order effect cancels exactly, giving the expression for $\Delta_{uv}^{Q}$ in Eq.~(\plaineqref{eq:query_pooling_gap_summary}). Its leading term couples query covariance with the curvature of the exact normalized block weight. The signed error can be positive or negative: a query centroid can either underestimate or overestimate the importance of a particular key block. In this analysis, $\Delta_{uv}^{Q}$ is the error relative to ground truth, not the difference between two approximate estimators, because no key-side approximation is used.

\paragraph{Dependence of the Hessian on all key blocks.}
To make the curvature term explicit without introducing key centroids, we write the exact token scores as
\begin{equation}
    z_j(\mathbf{q})=\mathbf{a}_j^{\top}\mathbf{q},
    \qquad
    \mathbf{a}_j=\frac{\mathbf{k}_j}{\sqrt d}.
    \label{eq:app_exact_token_scores}
\end{equation}
Here, $\mathbf{a}_j$ is the scaled original key, and $z_j(\mathbf{q})$ is its scaled QK score for query $\mathbf{q}$.
The corresponding attention-weighted statistics over all original keys are
\begin{equation}
    \overline{\mathbf{a}}(\mathbf{q})=\sum_{j=1}^{L}p_j(\mathbf{q})\mathbf{a}_j,
    \qquad
    \mathbf{C}_a(\mathbf{q})=\sum_{j=1}^{L}p_j(\mathbf{q})
    (\mathbf{a}_j-\overline{\mathbf{a}}(\mathbf{q}))
    (\mathbf{a}_j-\overline{\mathbf{a}}(\mathbf{q}))^{\top}.
    \label{eq:app_softmax_key_moments}
\end{equation}
Here, $\overline{\mathbf{a}}(\mathbf{q})$ and $\mathbf{C}_a(\mathbf{q})$ are the attention-weighted mean and covariance of the scaled keys, respectively.
Both are computed using the exact token attention weights $p_j(\mathbf{q})$ defined in Eq.~(\plaineqref{eq:app_exact_block_weight}).
Differentiating token-level Softmax gives
\begin{equation}
    \nabla_{\mathbf{q}}p_j(\mathbf{q})
    =p_j(\mathbf{q})(\mathbf{a}_j-\overline{\mathbf{a}}(\mathbf{q})),
    \qquad
    \nabla_{\mathbf{q}}\overline{\mathbf{a}}(\mathbf{q})
    =\mathbf{C}_a(\mathbf{q}).
    \label{eq:app_softmax_gradient_general}
\end{equation}
The second identity follows by substituting the first into the derivative of $\sum_j p_j(\mathbf{q})\mathbf{a}_j$. Here, $\nabla_{\mathbf{q}}\overline{\mathbf{a}}$ denotes its Jacobian.
Differentiating the token weight once more yields
\begin{equation}
    \nabla_{\mathbf{q}}^2p_j(\mathbf{q})
    =p_j(\mathbf{q})\left[
    (\mathbf{a}_j-\overline{\mathbf{a}}(\mathbf{q}))
    (\mathbf{a}_j-\overline{\mathbf{a}}(\mathbf{q}))^{\top}
    -\mathbf{C}_a(\mathbf{q})\right].
    \label{eq:app_token_softmax_hessian}
\end{equation}
Since $\phi_v$ sums the token weights in key block $\mathcal{K}_v$, its gradient and Hessian at the query centroid are
\begin{equation}
    \begin{aligned}
    \mathbf{g}_v
    &=\sum_{j\in\mathcal{K}_v}p_{uj}(\mathbf{a}_j-\overline{\mathbf{a}}_u),\\
    \mathbf{H}_v
    &=\sum_{j\in\mathcal{K}_v}p_{uj}\left[
    (\mathbf{a}_j-\overline{\mathbf{a}}_u)
    (\mathbf{a}_j-\overline{\mathbf{a}}_u)^{\top}
    -\mathbf{C}_{a,u}\right].
    \end{aligned}
    \label{eq:app_softmax_hessian}
\end{equation}
Here, $p_{uj}=p_j(\overline{\mathbf{q}}_u)$ is the attention weight assigned to key $j$ by the query centroid.
The quantities $\overline{\mathbf{a}}_u=\overline{\mathbf{a}}(\overline{\mathbf{q}}_u)$ and $\mathbf{C}_{a,u}=\mathbf{C}_a(\overline{\mathbf{q}}_u)$ are the mean and covariance of the scaled keys, respectively, weighted by these attention weights.
These expressions use individual keys, not key centroids.
Both $\overline{\mathbf{a}}_u$ and $\mathbf{C}_{a,u}$ depend on all original keys and their Softmax weights, including keys outside block $v$.
Consequently, the Hessian captures how the exact attention weight of block $v$ changes as the query changes, accounting for competition with all other keys under Softmax normalization.

\paragraph{Implications for query-centroid estimation.}
Accurate query-centroid estimation requires a small $|\Delta_{uv}^{Q}|$.
Since the covariance matrix $\boldsymbol{\Sigma}_{Q,u}$ is positive semidefinite, the leading error term satisfies
\begin{equation}
    \left|\frac{1}{2}\langle\mathbf{H}_v,\boldsymbol{\Sigma}_{Q,u}\rangle_F\right|
    \leq \frac{1}{2}\|\mathbf{H}_v\|_{\mathrm{op}}
    \operatorname{tr}(\boldsymbol{\Sigma}_{Q,u}),
    \qquad
    \operatorname{tr}(\boldsymbol{\Sigma}_{Q,u})
    =\frac{1}{n_u^Q}\sum_{i\in\mathcal{Q}_u}
    \|\boldsymbol{\epsilon}_i\|_2^2.
    \label{eq:app_query_leading_error_bound}
\end{equation}
Here, $\mathbf{H}_v=\mathbf{H}_{\phi_v}(\overline{\mathbf{q}}_u)$ is the Hessian at the query centroid, and $\|\mathbf{H}_v\|_{\mathrm{op}}$ is its operator norm, equal to its largest absolute eigenvalue.
The trace $\operatorname{tr}(\boldsymbol{\Sigma}_{Q,u})$ is the total within-block query variance, or the average squared Euclidean distance from the queries to their centroid.
The inequality follows by applying $|\boldsymbol{\epsilon}_i^\top\mathbf{H}_v\boldsymbol{\epsilon}_i|\leq\|\mathbf{H}_v\|_{\mathrm{op}}\|\boldsymbol{\epsilon}_i\|_2^2$ to each query and averaging.
Thus, a small product of the Hessian norm and the total query variance is sufficient for a small leading error.
Combining this inequality with Eq.~(\plaineqref{eq:app_query_remainder_bound}) also bounds the full approximation error:
\begin{equation}
    |\Delta_{uv}^{Q}|
    \leq\frac{1}{2}\|\mathbf{H}_v\|_{\mathrm{op}}
    \operatorname{tr}(\boldsymbol{\Sigma}_{Q,u})
    +\frac{M_v}{6n_u^Q}\sum_{i\in\mathcal{Q}_u}
    \|\boldsymbol{\epsilon}_i\|_2^3.
    \label{eq:app_query_full_error_bound}
\end{equation}
Controlling both terms on the right-hand side is sufficient to make the full error small.

If all queries in a block are identical, every deviation $\boldsymbol{\epsilon}_i$ is zero and query-centroid estimation is exact.
More generally, with bounded derivatives, both the second-order term and the remainder vanish as the query deviations approach zero.
In practice, clustered queries may still have different attention preferences.
Standard euclidean query clustering reduces but does not generally eliminate within-block query variance and does not account for the Hessian, so it does not by itself ensure that the product in Eq.~(\plaineqref{eq:app_query_leading_error_bound}) is sufficiently small.
We next evaluate the approximation error empirically under practical clustering settings.
Table~\ref{tab:query_representation_errors} shows that query-centroid estimation with full keys incurs mean attention-weight TV errors of $0.1335$ and $0.1667$ on Wan2.1-14B and HunyuanVideo, respectively, exceeding the errors obtained by retaining original queries with key centroids.

\paragraph{Relation to estimation with key centroids.}
The analysis above retains all original keys to examine the error introduced by query centroids.
PAMA additionally uses key centroids, which introduce a separate approximation error.
Figure~\ref{fig:retrieval_mismatches}(c) shows that retaining original queries reduces attention-weight approximation error when both estimates use the same key centroids.
Details of this experiment are provided in Appendix~\ref{app:mismatch_experiments}.
Table~\ref{tab:query_key_compression} compares different query and key compression settings and shows that retaining full queries while using key centroids achieves the highest recall among the compressed representations.
Together, these empirical results support preserving original queries for practical block retrieval and motivate PAMA.

\subsection{Derivation of the Query-Response Key Metric}
\label{app:key_metric_derivation}

For each key block, we decompose its keys around their mean:
\begin{equation}
    \mathbf{c}_v
    =\frac{1}{n_v^K}\sum_{j\in\mathcal{K}_v}\mathbf{k}_j,
    \qquad
    \boldsymbol{\delta}_{vj}
    =\mathbf{k}_j-\mathbf{c}_v.
    \label{eq:app_key_residual}
\end{equation}
Here, $\mathbf{c}_v$ is the centroid of the $n_v^K$ keys in $\mathcal{K}_v$, and $\boldsymbol{\delta}_{vj}$ is the deviation of key $j$ from this centroid.
As on the query side, the deviations have zero mean:
\begin{equation}
    \frac{1}{n_v^K}\sum_{j\in\mathcal{K}_v}\boldsymbol{\delta}_{vj}
    =\frac{1}{n_v^K}\sum_{j\in\mathcal{K}_v}\mathbf{k}_j-\mathbf{c}_v
    =\mathbf{0}.
    \label{eq:app_key_zero_mean}
\end{equation}
The corresponding scaled score deviation is
\begin{equation}
 x_{ivj}=\frac{\mathbf{q}_i^\top\boldsymbol{\delta}_{vj}}{\sqrt d}.
\end{equation}
Here, $x_{ivj}$ is the difference between the scaled QK score of query $\mathbf{q}_i$ and key $\mathbf{k}_j$ and the score obtained by replacing $\mathbf{k}_j$ with its block centroid $\mathbf{c}_v$.
Equation~(\plaineqref{eq:app_key_zero_mean}) immediately implies
\begin{equation}
    \frac{1}{n_v^K}\sum_{j\in\mathcal{K}_v}x_{ivj}
    =\frac{\mathbf{q}_i^{\top}}{\sqrt{d}}
    \left(\frac{1}{n_v^K}\sum_{j\in\mathcal{K}_v}\boldsymbol{\delta}_{vj}\right)
    =0.
    \label{eq:app_score_deviation_zero_mean}
\end{equation}

\paragraph{Exact decomposition.}
Using $\mathbf{k}_j=\mathbf{c}_v+\boldsymbol{\delta}_{vj}$, the exact block log-mass decomposes as
\begin{equation}
    \begin{aligned}
        S_{iv}^{\star}
        &=\log\sum_{j\in\mathcal{K}_v}
        \exp\!\left(\frac{\mathbf{q}_i^{\top}\mathbf{k}_j}{\sqrt{d}}\right)\\
        &=\log n_v^K+\frac{\mathbf{q}_i^{\top}\mathbf{c}_v}{\sqrt{d}}
        +\underbrace{\log\!\left[
        \frac{1}{n_v^K}\sum_{j\in\mathcal{K}_v}\exp(x_{ivj})
        \right]}_{\Delta_{iv}^{K}}.
    \end{aligned}
    \label{eq:app_key_exact_decomposition}
\end{equation}
Here, $S_{iv}^{\star}$ is the exact log-mass in Eq.~(\plaineqref{eq:key_logmass_summary}). 
The first two terms form $\widehat{S}_{iv}=\log n_v^K+\mathbf{q}_i^\top\mathbf{c}_v/\sqrt d$, the estimate in Eq.~(\plaineqref{eq:pama_block_score}).
Hence, $\Delta_{iv}^{K}=S_{iv}^{\star}-\widehat{S}_{iv}$ is the full log-mass approximation error.
Their relation to attention is
\begin{equation}
 P_{iv}^{\star}=\frac{e^{S_{iv}^{\star}}}{\sum_r e^{S_{ir}^{\star}}},\qquad
 \widehat P_{iv}=\frac{e^{\widehat S_{iv}}}{\sum_r e^{\widehat S_{ir}}}.
\end{equation}
Here, $P_{iv}^{\star}$ and $\widehat{P}_{iv}$ are the exact and estimated attention weights for query $i$, respectively, and the sums run over all nonempty key blocks.
These expressions show how key-centroid approximation affects attention weights through the block log-masses.
The term $\log n_v^K$ accounts for the number of keys represented by each centroid and is necessary when key blocks have different sizes.

\paragraph{Second-order expansion.}
To derive the second-order term and bound the remainder, we compute
\begin{equation}
    m_{2,iv}=\frac{1}{n_v^K}\sum_{j\in\mathcal{K}_v}x_{ivj}^{2},
    \qquad
    m_{3,iv}=\frac{1}{n_v^K}\sum_{j\in\mathcal{K}_v}|x_{ivj}|^{3}.
    \label{eq:app_score_deviation_moments}
\end{equation}
Here, $m_{2,iv}$ is the second moment and $m_{3,iv}$ is the absolute third moment of the centered scaled scores.
For each deviation, Taylor's theorem gives $\exp(x)=1+x+x^2/2+r_3(x)$.
Here, $r_3(x)$ is the remainder after retaining the terms $1$, $x$, and $x^2/2$.
If $|x_{ivj}|\leq\eta$ within the block, then $|r_3(x_{ivj})|\leq e^{\eta}|x_{ivj}|^3/6$.
Here, $\eta$ is an upper bound on the absolute scaled-score deviations for query $i$ within key block $\mathcal{K}_v$, satisfying $\eta\geq\max_{j\in\mathcal{K}_v}|x_{ivj}|$.
Averaging this expansion and using Eq.~(\plaineqref{eq:app_score_deviation_zero_mean}) therefore gives
\begin{equation}
    \frac{1}{n_v^K}\sum_{j\in\mathcal{K}_v}\exp(x_{ivj})
    =1+\frac{1}{2}m_{2,iv}+\rho_{iv}^{\exp},
    \qquad
    |\rho_{iv}^{\exp}|\leq\frac{e^{\eta}}{6}m_{3,iv}.
    \label{eq:app_expansion_exp}
\end{equation}
Here, $\rho_{iv}^{\exp}=(1/n_v^K) \sum_{j\in\mathcal{K}_v}r_3(x_{ivj})$ is the average of the Taylor remainders within key block $\mathcal{K}_v$ for query $i$.
The key deviations also give
\begin{equation}
    \boldsymbol{\Sigma}_{K,v}
    =\frac{1}{n_v^K}\sum_{j\in\mathcal{K}_v}
    \boldsymbol{\delta}_{vj}\boldsymbol{\delta}_{vj}^{\top}
    \in\mathbb{R}^{d\times d}.
    \label{eq:app_key_covariance}
\end{equation}
Here, $\boldsymbol{\Sigma}_{K,v}$ is the within-block key covariance, normalized by the number of keys.
The second moment can now be written as a quadratic form:
\begin{equation}
    m_{2,iv}
    =\frac{1}{dn_v^K}\sum_{j\in\mathcal{K}_v}
    (\mathbf{q}_i^{\top}\boldsymbol{\delta}_{vj})^2
    =\frac{1}{d}\mathbf{q}_i^{\top}
    \boldsymbol{\Sigma}_{K,v}\mathbf{q}_i.
    \label{eq:app_score_second_moment}
\end{equation}
Substituting Eq.~(\plaineqref{eq:app_expansion_exp}) into the exact error expression in Eq.~(\plaineqref{eq:app_key_exact_decomposition}) gives
\begin{equation}
    \Delta_{iv}^{K}
    =\log\left(1+\frac{1}{2}m_{2,iv}+\rho_{iv}^{\exp}\right).
    \label{eq:app_key_log_expansion_input}
\end{equation}
For sufficiently small score deviations, $y=m_{2,iv}/2+\rho_{iv}^{\exp}$ is close to zero.
Using the Taylor expansion $\log(1+y)=y+\mathcal{O}(y^2)$ and collecting the higher-order terms into $\mathcal{R}_{iv}^{K}$, we obtain
\begin{equation}
    \Delta_{iv}^{K}
    =\frac{1}{2}m_{2,iv}
    +\mathcal{R}_{iv}^{K}
    =\frac{1}{2d}\mathbf{q}_i^{\top}
    \boldsymbol{\Sigma}_{K,v}\mathbf{q}_i
    +\mathcal{R}_{iv}^{K}.
    \label{eq:app_key_second_order}
\end{equation}
To match the main-text notation, the leading term and full error are
\begin{equation}
 \epsilon_{iv}^{K}=\frac{m_{2,iv}}{2}=\frac{1}{2d}\mathbf{q}_i^\top\boldsymbol{\Sigma}_{K,v}\mathbf{q}_i,
 \qquad \Delta_{iv}^{K}=\epsilon_{iv}^{K}+\mathcal R_{iv}^{K}.
\end{equation}
Here, $\epsilon_{iv}^{K}$ is the leading error term, while $\mathcal{R}_{iv}^{K}$ collects higher-order contributions from the exponential and logarithmic expansions.
In particular, $\epsilon_{iv}^{K}$ is one half of the within-block variance of scaled QK scores, or $1/(2d)$ times the variance of unscaled scores. 
It is not the full approximation error.
For bounded, sufficiently small score deviations, there is a constant $C_{\eta}>0$ such that
\begin{equation}
    |\mathcal{R}_{iv}^{K}|
    \leq C_{\eta}\left(m_{3,iv}+m_{2,iv}^{2}\right).
    \label{eq:app_key_remainder_bound}
\end{equation}
The first term in the bound comes from the exponential remainder, while the second arises from the quadratic remainder of $\log(1+y)$.
Independently of the small-deviation approximation, Jensen's inequality gives the exact result
\begin{equation}
    \Delta_{iv}^{K}
    =\log\!\left(\frac{1}{n_v^K}\sum_{j\in\mathcal{K}_v}e^{x_{ivj}}\right)
    \geq \frac{1}{n_v^K}\sum_{j\in\mathcal{K}_v}x_{ivj}=0.
    \label{eq:app_key_jensen_bound}
\end{equation}
Thus, the centroid estimate is always a lower bound on the exact block log-mass, and the leading error term $\epsilon_{iv}^{K}$ is proportional to the variance of the keys after projection onto the current query.

\paragraph{Query-induced clustering metric.}
To construct a key-clustering objective that accounts for all current queries, we use
\begin{equation}
    \mathbf{G}_{Q}
    =\frac{1}{L}\sum_{i=1}^{L}\mathbf{q}_i\mathbf{q}_i^{\top}
    =\frac{1}{L}\mathbf{Q}^{\top}\mathbf{Q}
    \in\mathbb{R}^{d\times d}.
    \label{eq:app_query_second_moment}
\end{equation}
Here, $\mathbf{G}_Q$ is the uncentered second-moment matrix of the $L$ current queries, stored as rows of $\mathbf Q$. Unlike a centered covariance, it also retains their mean direction.
Using $\mathbf{x}^{\top}\mathbf{A}\mathbf{x}=\operatorname{tr}(\mathbf{A}\mathbf{x}\mathbf{x}^{\top})$ and the cyclic property of the trace, we have
\begin{equation}
    \begin{aligned}
        \overline{\epsilon}_{v}^{K}
        &=\frac{1}{L}\sum_{i=1}^{L}
        \frac{1}{2d}\mathbf{q}_i^{\top}\boldsymbol{\Sigma}_{K,v}\mathbf{q}_i\\
        &=\frac{1}{2d}\operatorname{tr}(\mathbf{G}_{Q}\boldsymbol{\Sigma}_{K,v})\\
        &=\frac{1}{2dn_v^K}\sum_{j\in\mathcal{K}_v}
        \boldsymbol{\delta}_{vj}^{\top}\mathbf{G}_{Q}
        \boldsymbol{\delta}_{vj}.
    \end{aligned}
    \label{eq:app_average_key_residual}
\end{equation}
Here, $\overline{\epsilon}_v^K=(1/L)\sum_i\epsilon_{iv}^K$ is the leading error term averaged over all current queries. 
Equation~(\plaineqref{eq:app_average_key_residual}) connects $\overline{\epsilon}_v^K$ to a query-dependent clustering objective.
Multiplying by the block size and summing over key blocks gives the total token-weighted distortion
\begin{equation}
    \begin{aligned}
        \mathcal{J}_{\mathrm{QRKC}}
        &=\sum_{v=1}^{N_K}\sum_{j\in\mathcal{K}_v}
        \boldsymbol{\delta}_{vj}^{\top}\mathbf{G}_{Q}
        \boldsymbol{\delta}_{vj}
        =2d\sum_{v=1}^{N_K}n_v^K\overline{\epsilon}_{v}^{K}\\
        &=\frac{1}{L}\sum_{i=1}^{L}\sum_{v=1}^{N_K}
        \sum_{j\in\mathcal{K}_v}
        \left(\mathbf{q}_i^{\top}\boldsymbol{\delta}_{vj}\right)^2.
    \end{aligned}
    \label{eq:app_qrkc_exact_distortion}
\end{equation}
The last equality follows by substituting the definition of $\mathbf{G}_{Q}$ into each quadratic form.
$\mathcal{J}_{\mathrm{QRKC}}/d$ measures the squared error introduced by replacing each key with its block centroid in the scaled attention logits.
This error is summed over all keys and averaged over all queries.
This objective weights each key block's query-averaged leading error term $\overline{\epsilon}_v^K$ by its number of keys $n_v^K$.

Ordinary Euclidean clustering instead minimizes
\begin{equation}
    \mathcal{J}_{\mathrm{Euc}}
    =\sum_{v=1}^{N_K}\sum_{j\in\mathcal{K}_v}
    \boldsymbol{\delta}_{vj}^{\top}\mathbf{I}\boldsymbol{\delta}_{vj},
    \label{eq:app_euclidean_objective}
\end{equation}
where $\mathbf{I}$ is the identity matrix.
Comparing Eqs.~(\plaineqref{eq:app_qrkc_exact_distortion}) and (\plaineqref{eq:app_euclidean_objective}) shows that the mismatch is precisely the use of $\mathbf{I}$ instead of the query-induced metric $\mathbf{G}_{Q}$.

\paragraph{Euclidean realization.}
To optimize the query-dependent objective with a standard Euclidean K-means implementation, we need a key transformation whose squared Euclidean distances equal the quadratic forms defined by $\mathbf{G}_Q$.
Directly mapping each key to $\mathbf{Q}\mathbf{k}_j/\sqrt L$ satisfies this requirement, but produces an $L$-dimensional feature vector instead of a $d$-dimensional one.
For the long video sequences considered here, $L \gg d$, so clustering these vectors would increase both storage and distance-computation costs.
We instead factorize the $d\times d$ matrix $\mathbf{G}_Q$ to construct a transformation that retains the feature dimension $d$ while realizing the same distances.
We first establish this property using a symmetric square root and then show how Cholesky decomposition provides the factor used in our implementation.
The matrix $\mathbf{G}_{Q}$ is positive semidefinite because, for any $\mathbf{x}\in\mathbb{R}^{d}$,
\begin{equation}
    \mathbf{x}^{\top}\mathbf{G}_{Q}\mathbf{x}
    =\frac{1}{L}\sum_{i=1}^{L}(\mathbf{q}_i^{\top}\mathbf{x})^2
    =\frac{1}{L}\|\mathbf{Q}\mathbf{x}\|_2^2\geq0.
    \label{eq:app_query_metric_psd}
\end{equation}
It therefore admits the eigendecomposition and principal square root
\begin{equation}
    \mathbf{G}_{Q}=\mathbf{U}\boldsymbol{\Lambda}\mathbf{U}^{\top},
    \qquad
    \mathbf{G}_{Q}^{1/2}
    =\mathbf{U}\boldsymbol{\Lambda}^{1/2}\mathbf{U}^{\top},
    \label{eq:app_query_metric_sqrt}
\end{equation}
where $\mathbf{U}\in\mathbb{R}^{d\times d}$ contains orthonormal eigenvectors and $\boldsymbol{\Lambda}\in\mathbb{R}^{d\times d}$ is diagonal with nonnegative eigenvalues.
The symmetric-square-root transformation is
\begin{equation}
 \widetilde{\mathbf{k}}_j^{\mathrm{sqrt}}=\mathbf{G}_{Q}^{1/2}\mathbf{k}_j,\qquad
 \widetilde{\mathbf{c}}_v^{\mathrm{sqrt}}=\mathbf{G}_{Q}^{1/2}\mathbf{c}_v.
\end{equation}
Here, the superscript $\mathrm{sqrt}$ identifies the square-root realization of the transformed key and centroid.
For the residual $\boldsymbol{\delta}_{vj}=\mathbf{k}_j-\mathbf{c}_v$, we have
\begin{equation}
    \|\widetilde{\mathbf{k}}_j^{\mathrm{sqrt}}-
    \widetilde{\mathbf{c}}_v^{\mathrm{sqrt}}\|_2^2
    =\|\mathbf{G}_{Q}^{1/2}\boldsymbol{\delta}_{vj}\|_2^2
    =(\mathbf{k}_j-\mathbf{c}_v)^{\top}\mathbf{G}_{Q}
    (\mathbf{k}_j-\mathbf{c}_v).
    \label{eq:app_transformed_distance}
\end{equation}

\begin{figure}[H]
    \centering
    \includegraphics[width=0.85\linewidth]{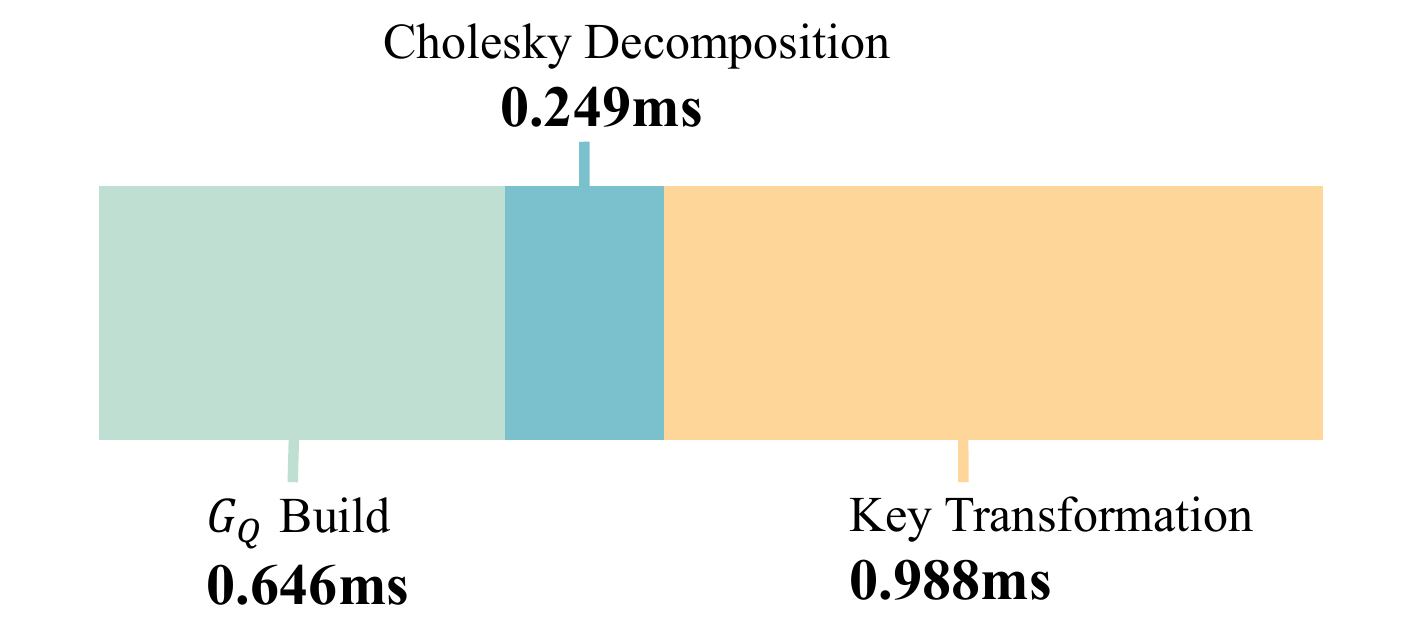}
    \vspace{-10pt}
    \caption{
        Time breakdown of QRKC overhead excluding K-means clustering. 
        The total measured overhead remains below 2\,ms.
        We use the same experimental settings as in
        Table~\ref{tab:pama_kernel_overhead}.
    }
    \label{fig:qrkc_breakdown}
\end{figure}

When $\mathbf{G}_{Q}$ is positive definite, it also admits the Cholesky decomposition
\begin{equation}
    \mathbf{G}_{Q}=\mathbf{R}\mathbf{R}^{\top}.
    \label{eq:app_query_metric_cholesky}
\end{equation}
The corresponding transformation is
\begin{equation}
 \widetilde{\mathbf{k}}_j^{\mathrm{chol}}=\mathbf{R}^{\top}\mathbf{k}_j,\qquad
 \widetilde{\mathbf{c}}_v^{\mathrm{chol}}=\mathbf{R}^{\top}\mathbf{c}_v.
\end{equation}
Here, the superscript $\mathrm{chol}$ identifies the Cholesky realization, whose transformed key is $\widetilde{\mathbf{k}}_j$ in Eq.~(\plaineqref{eq:qrkc_key_transform}).
Their squared Euclidean distance is
\begin{equation}
    \begin{aligned}
        \|\widetilde{\mathbf{k}}_j^{\mathrm{chol}}-
        \widetilde{\mathbf{c}}_v^{\mathrm{chol}}\|_2^2
        &=\|\mathbf{R}^{\top}\boldsymbol{\delta}_{vj}\|_2^2\\
        &=\boldsymbol{\delta}_{vj}^{\top}
        \mathbf{R}\mathbf{R}^{\top}\boldsymbol{\delta}_{vj}\\
        &=\boldsymbol{\delta}_{vj}^{\top}
        \mathbf{G}_{Q}\boldsymbol{\delta}_{vj}.
    \end{aligned}
    \label{eq:app_cholesky_transformed_distance}
\end{equation}
In particular, the direct query-based transformation and the Cholesky transformation satisfy
\begin{equation}
    \left\|\frac{\mathbf{Q}\boldsymbol{\delta}_{vj}}{\sqrt L}\right\|_2^2
    =\boldsymbol{\delta}_{vj}^{\top}\mathbf{G}_Q\boldsymbol{\delta}_{vj}
    =\|\mathbf{R}^{\top}\boldsymbol{\delta}_{vj}\|_2^2.
    \label{eq:app_direct_query_cholesky_equivalence}
\end{equation}
The vector on the left has $L$ components, whereas the transformed residual on the right has only $d$ components.
Thus, Cholesky decomposition lets us realize the query-dependent distance in Eq.~(\plaineqref{eq:qrkc_transformed_distance}) without constructing an $L$-dimensional representation for every key or modifying the distance computation in K-means.
For each attention head, we compute $\mathbf{R}$ once before K-means and use it to transform all keys.
For the same numbers of keys, clusters, and iterations, the K-means stage retains the computational complexity of clustering the original $d$-dimensional keys.
The additional work is computing $\mathbf{G}_Q$, factorizing it, and transforming the keys.
Figure~\ref{fig:qrkc_breakdown} shows the time breakdown of these operations, whose total measured overhead remains below 2\,ms across the evaluated settings.
Equations~(\plaineqref{eq:app_transformed_distance}) and
(\plaineqref{eq:app_cholesky_transformed_distance}) show that the symmetric-square-root and Cholesky transformations produce the same squared distance under $\mathbf{G}_{Q}$.
Because a linear transformation also maps each original centroid to the centroid of the transformed keys, summing either distance over all clusters gives
\begin{equation}
    \sum_{v=1}^{N_K}\sum_{j\in\mathcal{K}_v}
    \|\widetilde{\mathbf{k}}_j-\widetilde{\mathbf{c}}_v\|_2^2
    =\sum_{v=1}^{N_K}\sum_{j\in\mathcal{K}_v}
    \boldsymbol{\delta}_{vj}^{\top}\mathbf{G}_{Q}
    \boldsymbol{\delta}_{vj}
    =\mathcal{J}_{\mathrm{QRKC}}.
    \label{eq:app_euclidean_qrkc_equivalence}
\end{equation}
Therefore, Euclidean K-means after either transformation optimizes exactly the same QRKC objective in Eq.~(\plaineqref{eq:app_qrkc_exact_distortion}).
The symmetric square root remains valid when $\mathbf{G}_{Q}$ is rank deficient, whereas standard Cholesky decomposition requires a positive-definite matrix.

\paragraph{Numerically stable Cholesky factorization.}
To ensure a stable Cholesky factorization, we apply scale-aware shrinkage to the query-induced metric.
The scale used for stabilization is
\begin{equation}
    s_Q=\max\!\left(\frac{1}{d}\operatorname{tr}(\mathbf{G}_{Q}),\;s_{\min}\right),
    \qquad s_{\min}>0.
    \label{eq:app_query_metric_scale}
\end{equation}
Here, $s_Q$ is the average diagonal value floored at $s_{\min}>0$, which is set to $10^{-12}$ in the implementation.
We construct the stabilized metric
\begin{equation}
    \overline{\mathbf{G}}_{Q}
    =(1-\alpha)\mathbf{G}_{Q}
    +\alpha s_Q\mathbf{I}
    +\epsilon s_Q\mathbf{I},
    \label{eq:app_regularized_query_metric}
\end{equation}
where $\alpha\in[0,1]$ is the shrinkage coefficient and $\epsilon>0$ is a small stability constant.
The isotropic terms make $\overline{\mathbf{G}}_{Q}$ positive definite while keeping their scale consistent with $\mathbf{G}_{Q}$.
We then compute
\begin{equation}
    \overline{\mathbf{G}}_{Q}=\mathbf{R}\mathbf{R}^{\top}
    \label{eq:app_regularized_cholesky}
\end{equation}
and transform each key by $\widetilde{\mathbf{k}}_j=\mathbf{R}^{\top}\mathbf{k}_j$.
Following the same argument as Eq.~(\plaineqref{eq:app_cholesky_transformed_distance}), Euclidean K-means on these transformed keys optimizes the stabilized objective
\begin{equation}
    \overline{\mathcal{J}}_{\mathrm{QRKC}}
    =\sum_{v=1}^{N_K}\sum_{j\in\mathcal{K}_v}
    \boldsymbol{\delta}_{vj}^{\top}
    \overline{\mathbf{G}}_{Q}\boldsymbol{\delta}_{vj}.
    \label{eq:app_regularized_qrkc_objective}
\end{equation}
The stabilized objective satisfies
\begin{equation}
 \overline{\mathcal J}_{\mathrm{QRKC}}=(1-\alpha)\mathcal J_{\mathrm{QRKC}}+(\alpha+\epsilon)s_Q\mathcal J_{\mathrm{Euc}}.
\end{equation}
Thus, stabilization adds an isotropic penalty, so the resulting objective differs from the unregularized objective.
Even small stabilization terms can change cluster assignments when the original metric assigns very little weight to certain directions.
The scalar $\epsilon$ here is a numerical stability parameter, distinct from the leading approximation error $\epsilon_{iv}^{K}$.

K-means optimizes the specified objective iteratively. 
The shrinkage term also provides a controlled interpolation between the query-induced metric and ordinary Euclidean distance: setting $\alpha=0$ recovers the original metric up to the small stability term, whereas $\alpha=1$ yields an isotropic metric up to a global scaling factor.
With $\alpha=0.05$ and the numerical stabilization described above, we observed no Cholesky decomposition failures in our experiments.

\section{Fused PAMA Kernel}
\label{app:fused_pama_kernel}

Algorithms~\ref{alg:full_k_pama_kernel} and~\ref{alg:two_pass_pama_kernel} describe two fused implementations for one attention head.
Programs for different attention heads and query blocks are executed in parallel.
The queries are stored contiguously according to their query-block assignments, and $o_u$ and $o_{u+1}$ denote the beginning and end of query block $\mathcal{Q}_u$, respectively.
The Full-$K$ kernel retains the logits over all key centroids in registers and is used when $N_K\leq1024$.
For larger $N_K$, retaining the complete logit row creates excessive register pressure, so we use a two-pass kernel that processes the key centroids in tiles.

In both algorithms, $i$ indexes a query in the current tile $I$, and $v$ indexes a key block in the current key-centroid tile $J$ or in $\{1,\ldots,N_K\}$ for Full-$K$.
Tile entries $S_{iv}$ and $P_{iv}$ use these query and key-block indices.
All indexed assignments apply in parallel over the relevant indices in these ranges, not as serial loops.

\begin{algorithm}[H]
    \caption{Full-$K$ fused PAMA kernel}
    \label{alg:full_k_pama_kernel}
    \begin{algorithmic}[1]
        \Require Reordered queries $\mathbf{Q}\in\mathbb{R}^{L\times d}$; query-block offsets $\{o_u\}_{u=1}^{N_Q+1}$; key centroids $\mathbf{C}\in\mathbb{R}^{N_K\times d}$; key-block sizes $\{n_v^K\}_{v=1}^{N_K}$
        \Ensure Block importance $\widehat{\mathbf{W}}^{\mathrm{PAMA}}\in\mathbb{R}^{N_Q\times N_K}$
        \ForAll{query blocks $u=1,\ldots,N_Q$ in parallel}
            \State $\mathbf{a}\gets\mathbf{0}\in\mathbb{R}^{N_K}$ \Comment{register-resident accumulator}
            \For{each query tile $I\subseteq\{o_u,\ldots,o_{u+1}-1\}$}
                \State $\mathbf{S}\gets\mathbf{0}\in\mathbb{R}^{|I|\times N_K}$ \Comment{register-resident logits}
                \For{each feature tile $D_t\subseteq\{1,\ldots,d\}$}
                    \State Load $\mathbf{Q}_{I,D_t}$ and $\mathbf{C}_{:,D_t}$
                    \State $\mathbf{S}\gets\mathbf{S}+\mathbf{Q}_{I,D_t}\mathbf{C}_{:,D_t}^{\top}$
                \EndFor
                \State $S_{iv}\gets S_{iv}/\sqrt{d}+\log n_v^K$
                \State $m_i\gets\max_v S_{iv}$ \Comment{row maximum}
                \State $E_{iv}\gets\exp(S_{iv}-m_i)$ \Comment{register resident}
                \State $z_i\gets\sum_v E_{iv}$ \Comment{row normalization factor}
                \State $P_{iv}\gets E_{iv}/z_i$ \Comment{row-wise Softmax}
                \State $a_v\gets a_v+\sum_{i\in I}P_{iv}$ \Comment{sum over queries}
            \EndFor
            \State $\widehat{\mathbf{W}}_{u,:}^{\mathrm{PAMA}}\gets\mathbf{a}/(o_{u+1}-o_u)$ \Comment{the only HBM write}
        \EndFor
        \State \Return $\widehat{\mathbf{W}}^{\mathrm{PAMA}}$
    \end{algorithmic}
\end{algorithm}

The intermediate tensors and attention-weight accumulator in Algorithm~\ref{alg:full_k_pama_kernel} remain in on-chip registers, without materializing the full $L\times N_K$ logit or probability matrix in HBM.
It computes every QK-centroid product once, but its register footprint grows with $N_K$.

\begin{algorithm}[H]
    \caption{Two-pass fused PAMA kernel}
    \label{alg:two_pass_pama_kernel}
    \begin{algorithmic}[1]
        \Require Reordered queries $\mathbf{Q}\in\mathbb{R}^{L\times d}$; query-block offsets $\{o_u\}_{u=1}^{N_Q+1}$; key centroids $\mathbf{C}\in\mathbb{R}^{N_K\times d}$; key-block sizes $\{n_v^K\}_{v=1}^{N_K}$
        \Ensure Block importance $\widehat{\mathbf{W}}^{\mathrm{PAMA}}\in\mathbb{R}^{N_Q\times N_K}$
        \State Initialize $\widehat{\mathbf{W}}^{\mathrm{PAMA}}\gets\mathbf{0}$ in HBM
        \ForAll{query blocks $u=1,\ldots,N_Q$ in parallel}
            \For{each query tile $I\subseteq\{o_u,\ldots,o_{u+1}-1\}$}
                \State Load $\mathbf{Q}_{I,:}$
                \State $m_i\gets-\infty$, $z_i\gets0$ \Comment{Softmax states in registers}
                \For{each key-centroid tile $J\subseteq\{1,\ldots,N_K\}$} \Comment{Pass 1}
                    \State $\mathbf{S}\gets\mathbf{Q}_{I,:}\mathbf{C}_{J,:}^{\top}/\sqrt{d}$
                    \State $S_{iv}\gets S_{iv}+\log n_v^K$
                    \State $m_i'\gets\max(m_i,\max_{v\in J}S_{iv})$
                    \State $z_i\gets z_i\exp(m_i-m_i')+\sum_{v\in J}\exp(S_{iv}-m_i')$
                    \State $m_i\gets m_i'$
                \EndFor
                \State $\ell_i\gets m_i+\log z_i$ \Comment{row-wise LogSumExp in registers}
                \For{each key-centroid tile $J\subseteq\{1,\ldots,N_K\}$} \Comment{Pass 2}
                    \State Recompute $\mathbf{S}\gets\mathbf{Q}_{I,:}\mathbf{C}_{J,:}^{\top}/\sqrt{d}$
                    \State $S_{iv}\gets S_{iv}+\log n_v^K$
                    \State $P_{iv}\gets\exp(S_{iv}-\ell_i)$ \Comment{in registers}
                    \State $\widehat{W}_{uv}^{\mathrm{PAMA}}\mathrel{+}=\sum_{i\in I}P_{iv}/(o_{u+1}-o_u)$ \Comment{HBM accumulation}
                \EndFor
            \EndFor
        \EndFor
        \State \Return $\widehat{\mathbf{W}}^{\mathrm{PAMA}}$
    \end{algorithmic}
\end{algorithm}
\vspace{-10pt}
The first pass computes row-wise LogSumExp through online updates over key-centroid tiles.
The second recomputes the logits, normalizes each tile, and accumulates block importance.
This approximately doubles QK-centroid GEMM work but keeps register usage independent of $N_K$ for fixed tile sizes.
Logits, probabilities, and LogSumExp values remain on chip rather than in HBM, allowing larger $N_K$ than Full-$K$.

Table~\ref{tab:pama_kernel_overhead} compares PyTorch with the Two-pass and Full-$K$ kernels in Triton (\cite{triton}) and TileLang (\cite{tilelang}).
Both benchmarks use batch size 1, $N_Q=256$, $N_K=1{,}024$, and $d=128$, with $(H,L)=(24,115{,}200)$ for HunyuanVideo and $(40,75{,}600)$ for Wan2.1-14B.

Additional full-model timing shows that Triton Full-$K$ completes PAMA block-importance estimation across all layers in less than 1\,s on both models.
Four such block-importance recomputations during video generation account for less than 0.21\% of the end-to-end latencies in Table~\ref{tab:main_quality_efficiency}.
\vspace{-5pt}
\begin{table}[!htbp]
    \centering
    \caption{PAMA runtime and additional GPU memory. Speedup, memory reduction (multiplicative factor), and memory saving are relative to PyTorch.}
    \label{tab:pama_kernel_overhead}
    \small
    \setlength{\tabcolsep}{3.6pt}
    \renewcommand{\arraystretch}{1.08}
    \begin{tabularx}{\linewidth}{@{}ll*{5}{>{\centering\arraybackslash}X}@{}}
        \toprule
        \multirow{2}{*}{Model} & \multirow{2}{*}{Implementation} & \multicolumn{2}{c}{Runtime} & \multicolumn{3}{c}{Additional GPU Memory} \\
        \cmidrule(lr){3-4}\cmidrule(l){5-7}
        & & Latency $\downarrow$ & Speedup $\uparrow$ & Peak $\downarrow$ & Reduction $\uparrow$ & Saving $\uparrow$ \\
        \midrule
        \multirow{5}{*}{HunyuanVideo}
        & PyTorch                & 87.95ms & 1.00$\times$ & 21.09 GB & 1.00$\times$ & 0.00\% \\
        & Triton (Two-pass)      & 28.12ms & 3.13$\times$ & 24.12 MB & 895.60$\times$  & 99.89\%   \\
        & Triton (Full-$K$)      & 15.82ms & 5.56$\times$ & 24.17 MB & 893.86$\times$  & 99.89\%   \\
        & TileLang (Two-pass)    & 21.35ms & 4.12$\times$ & 24.12 MB & 895.60$\times$  & 99.89\%   \\
        & TileLang (Full-$K$)    & 21.08ms & 4.17$\times$ & 24.17 MB & 893.86$\times$  & 99.89\%   \\
        \midrule
        \multirow{5}{*}{Wan2.1-14B}
        & PyTorch                & 92.81ms & 1.00$\times$ & 23.07 GB & 1.00$\times$ & 0.00\% \\
        & Triton (Two-pass)      & 31.35ms & 2.96$\times$ & 40.20 MB & 587.71$\times$   & 99.83\%   \\
        & Triton (Full-$K$)      & 17.72ms & 5.24$\times$ & 40.27 MB & 586.57$\times$   & 99.83\%   \\
        & TileLang (Two-pass)    & 23.82ms & 3.90$\times$ & 40.20 MB & 587.71$\times$   & 99.83\%   \\
        & TileLang (Full-$K$)    & 23.22ms & 4.00$\times$ & 40.27 MB & 586.57$\times$   & 99.83\%   \\
        \bottomrule
    \end{tabularx}
\end{table}

\section{QRKC Pseudocode}
\label{app:qrkc_pseudocode}

Algorithm~\ref{alg:qrkc} summarizes QRKC for one attention head.
The transformed keys are used only to determine the cluster assignments. The labels are then applied to the original keys and values, and PAMA uses centroids computed in the original key space.

\begin{algorithm}[H]
    \caption{Query-Response Key Clustering (QRKC)}
    \label{alg:qrkc}
    \begin{algorithmic}[1]
        \Require Queries $\mathbf{Q}\in\mathbb{R}^{L\times d}$; keys $\mathbf{K}\in\mathbb{R}^{L\times d}$; values $\mathbf{V}\in\mathbb{R}^{L\times d}$; number of key clusters $N_K$; K-means iterations $T$; shrinkage $\alpha$; stability constant $\epsilon$
        \Ensure Key labels $\mathbf{z}$; permutation $\boldsymbol{\pi}_K$; reordered keys $\mathbf{K}'$ and values $\mathbf{V}'$; original-space centroids $\{\mathbf{c}_v\}_{v=1}^{N_K}$; cluster sizes $\{n_v^K\}_{v=1}^{N_K}$
        \State $\mathbf{G}_{Q}\gets\mathbf{Q}^{\top}\mathbf{Q}/L$
        \State $\mathbf{G}_{Q}\gets(\mathbf{G}_{Q}+\mathbf{G}_{Q}^{\top})/2$ \Comment{enforce numerical symmetry}
        \State $s_Q\gets\max\!\left(\operatorname{tr}(\mathbf{G}_{Q})/d,10^{-12}\right)$
        \State $\overline{\mathbf{G}}_{Q}\gets(1-\alpha)\mathbf{G}_{Q}+\alpha s_Q\mathbf{I}+\epsilon s_Q\mathbf{I}$
        \State $\mathbf{R}\gets\operatorname{Cholesky}(\overline{\mathbf{G}}_{Q})$ \Comment{$\overline{\mathbf{G}}_{Q}=\mathbf{R}\mathbf{R}^{\top}$}
        \State $\widetilde{\mathbf{K}}\gets\mathbf{K}\mathbf{R}$ \Comment{query-conditioned key features}
        \State Initialize transformed centroids $\{\widetilde{\boldsymbol{\mu}}_v\}_{v=1}^{N_K}$ from $\widetilde{\mathbf{K}}$
        \For{$t=1,\ldots,T$}
            \ForAll{keys $j=1,\ldots,L$ in parallel}
                \State $z_j\gets\arg\min_{v\in\{1,\ldots,N_K\}}\|\widetilde{\mathbf{k}}_j-\widetilde{\boldsymbol{\mu}}_v\|_2^2$
            \EndFor
            \ForAll{clusters $v=1,\ldots,N_K$ in parallel}
                \State $\widetilde{\boldsymbol{\mu}}_v\gets\operatorname{mean}\{\widetilde{\mathbf{k}}_j:z_j=v\}$
            \EndFor
        \EndFor
        \State $\boldsymbol{\pi}_K\gets\operatorname{argsort}(\mathbf{z})$
        \State $\mathbf{K}'\gets\mathbf{K}[\boldsymbol{\pi}_K,:]$, $\mathbf{V}'\gets\mathbf{V}[\boldsymbol{\pi}_K,:]$
        \ForAll{clusters $v=1,\ldots,N_K$ in parallel}
            \State $\mathcal{K}_v\gets\{j:z_j=v\}$, $n_v^K\gets|\mathcal{K}_v|$
            \State $\mathbf{c}_v\gets(n_v^K)^{-1}\sum_{j\in\mathcal{K}_v}\mathbf{k}_j$ \Comment{centroid in the original key space}
        \EndFor
        \State \Return $\mathbf{z},\boldsymbol{\pi}_K,\mathbf{K}',\mathbf{V}',\{\mathbf{c}_v,n_v^K\}_{v=1}^{N_K}$
    \end{algorithmic}
\end{algorithm}

Because $\widetilde{\mathbf{K}}=\mathbf{K}\mathbf{R}$, Euclidean K-means in the transformed space minimizes the stabilized QRKC objective in Eq.~(\plaineqref{eq:app_regularized_qrkc_objective}).
The transformation does not change the feature dimension, and no transformed values are required.
Thus, K-means retains the same computational complexity for the same token count, cluster count, and number of iterations.

\vspace{-5pt}

\newpage

\section{Sensitivity to Query and Key Centroid Counts}
\label{app:centroid_sensitivity}

We analyze the sensitivity of PARK to the numbers of query and key centroids, denoted by $N_Q$ and $N_K$, respectively.

\begin{table}[H]
    \centering
    \caption{Sensitivity to the numbers of query and key centroids on HunyuanVideo.}
    \label{tab:centroid_count_sensitivity_hy}
    \small
    \setlength{\tabcolsep}{4.2pt}
    \renewcommand{\arraystretch}{1.08}
    \begin{tabularx}{\linewidth}{@{}*{7}{>{\centering\arraybackslash}X}@{}}
        \toprule
        $N_Q$ & $N_K$ & PSNR $\uparrow$ & SSIM $\uparrow$ & LPIPS $\downarrow$ & Latency $\downarrow$ & Speedup $\uparrow$ \\
        \midrule
        64  & 1,024 & 29.38 & 0.926 & 0.175 & 1817s & 1.68$\times$ \\
        128 & 1,024 & 29.85 & 0.931 & 0.168 & 1827s & 1.67$\times$ \\
        192 & 1,024 & 29.94 & 0.933 & 0.167 & 1841s & 1.66$\times$ \\
        256 & 1,024 & 30.13 & 0.934 & 0.164 & 1855s & 1.64$\times$ \\
        320 & 1,024 & 30.13 & 0.935 & 0.162 & 1875s & 1.63$\times$ \\
        \midrule
        128 & 512   & 29.56 & 0.928 & 0.174 & 1830s & 1.67$\times$ \\
        128 & 768   & 29.74 & 0.931 & 0.170 & 1834s & 1.66$\times$ \\
        \bottomrule
    \end{tabularx}
\end{table}

\begin{table}[H]
    \centering
    \caption{Sensitivity to the numbers of query and key centroids on Wan2.1-1.3B.}
    \label{tab:centroid_count_sensitivity_wan}
    \small
    \setlength{\tabcolsep}{4.2pt}
    \renewcommand{\arraystretch}{1.08}
    \begin{tabularx}{\linewidth}{@{}*{7}{>{\centering\arraybackslash}X}@{}}
        \toprule
        $N_Q$ & $N_K$ & PSNR $\uparrow$ & SSIM $\uparrow$ & LPIPS $\downarrow$ & Latency $\downarrow$ & Speedup $\uparrow$ \\
        \midrule
        64  & 512 & 27.30 & 0.903 & 0.183 & 410s & 1.82$\times$ \\
        128 & 512 & 27.52 & 0.908 & 0.177 & 423s & 1.76$\times$ \\
        192 & 512 & 27.66 & 0.911 & 0.174 & 427s & 1.74$\times$ \\
        256 & 512 & 27.71 & 0.911 & 0.173 & 430s & 1.73$\times$ \\
        320 & 512 & 27.79 & 0.913 & 0.172 & 440s & 1.69$\times$ \\
        \midrule
        128 & 768   & 27.67 & 0.910 & 0.175 & 416s & 1.79$\times$ \\
        128 & 1024  & 27.72 & 0.912 & 0.173 & 420s & 1.77$\times$ \\
        \bottomrule
    \end{tabularx}
\end{table}

\section{Clustering and Sparse-Mask Reuse}
\label{app:reuse_schedule}

We evaluate the effect of the recomputation schedule on HunyuanVideo with 50 denoising steps.
Let $\mathcal{T}$ denote the set of denoising step indices at which clustering results and sparse masks are computed or recomputed. Both are reused during subsequent sparse-attention steps until the next recomputation.
Table~\ref{tab:reuse_schedule_hy} compares four schedules in terms of generation quality and inference efficiency.

\begin{table}[H]
    \centering
    \caption{Effect of clustering and sparse-mask recomputation schedules on HunyuanVideo with 50 denoising steps. Speedup is measured relative to dense attention.}
    \label{tab:reuse_schedule_hy}
    \small
    \setlength{\tabcolsep}{4.2pt}
    \renewcommand{\arraystretch}{1.08}
    \begin{tabularx}{\linewidth}{@{}c*{5}{>{\centering\arraybackslash}X}@{}}
        \toprule
        $\mathcal{T}$ & PSNR $\uparrow$ & SSIM $\uparrow$ & LPIPS $\downarrow$ & Latency $\downarrow$ & Speedup $\uparrow$ \\
        \midrule
        $\{10\}$             & 29.55 & 0.928 & 0.174 & 1829s & 1.67$\times$ \\
        $\{10,30\}$          & 29.67 & 0.930 & 0.171 & 1819s & 1.68$\times$ \\
        $\{10,20,30\}$       & 29.84 & 0.932 & 0.168 & 1819s & 1.68$\times$ \\
        $\{10,20,30,40\}$    & 29.85 & 0.931 & 0.168 & 1827s & 1.67$\times$ \\
        \bottomrule
    \end{tabularx}
\end{table}

As shown in Table~\ref{tab:reuse_schedule_hy}, increasing the number of recomputations generally improves similarity to dense attention, while three and four recomputations yield nearly identical results.
All four schedules achieve similar end-to-end latency.
These results support reusing clustering results and sparse masks across diffusion steps, with occasional recomputation to preserve generation quality.

\newpage

\section{Efficiency-Quality Trade-off Study}

\begin{figure}[H]
    \centering
    \includegraphics[width=\linewidth]{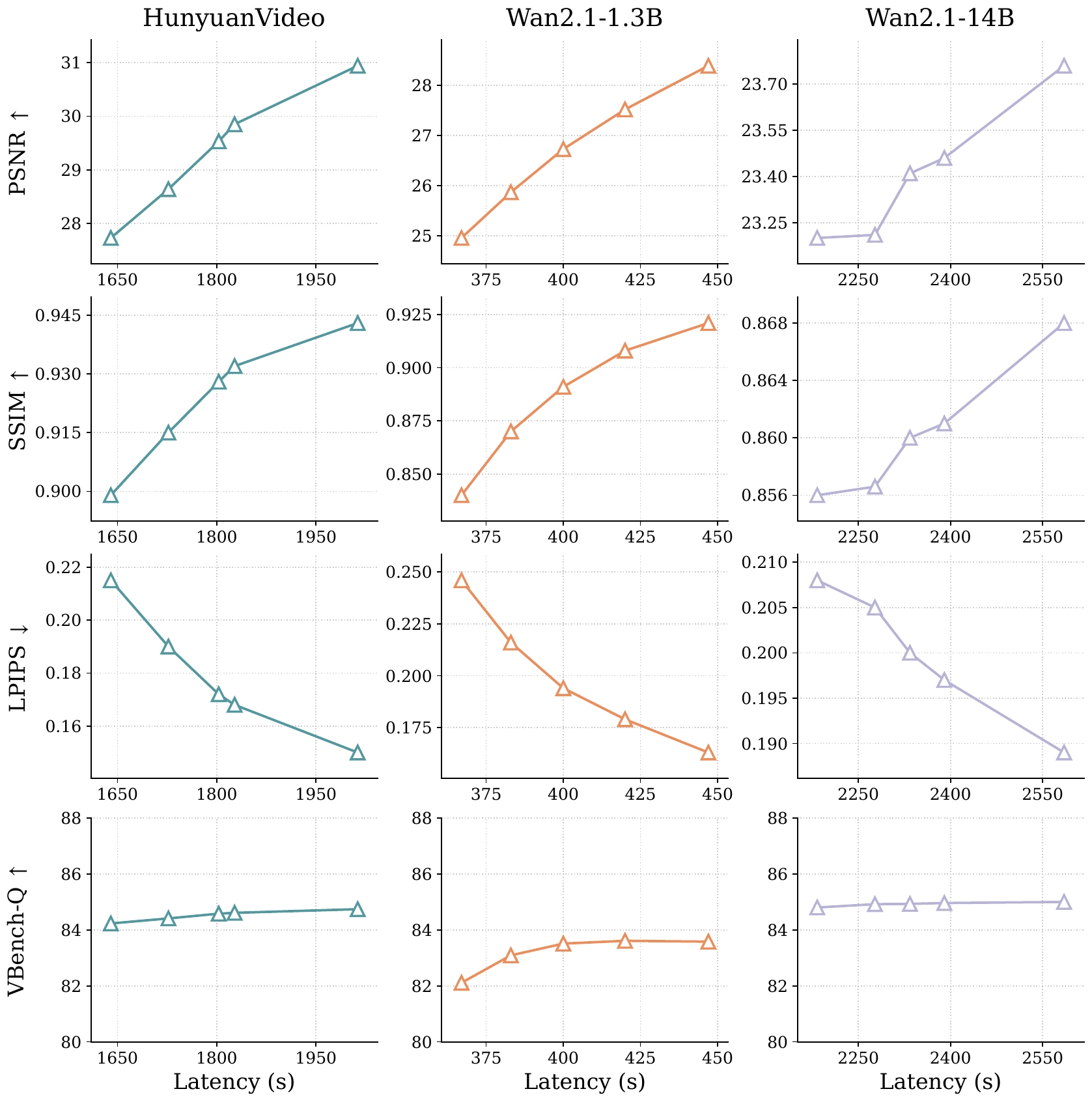}
    \caption{
        Efficiency-quality trade-off of PARK.
    }
    \label{fig:quality_efficiency_curve}
\end{figure}

We study the efficiency-quality trade-off of PARK under different sparsity settings on a subset of VBench prompts using HunyuanVideo, Wan2.1-1.3B, and Wan2.1-14B.
As shown in Figure~\ref{fig:quality_efficiency_curve}, although similarity to dense attention decreases as latency is reduced, PARK maintains high generation quality across the tested sparsity settings, as reflected by its relatively stable VBench-Q scores.

\newpage

\section{Query and Key Compression under Top-k Retrieval}
\label{app:topk_compression}

We extend the query/key compression comparison in Table~\ref{tab:query_key_compression} to Top-k retrieval with a fixed retrieval ratio $\rho$.
For each query block, we retrieve the top $\rho$ fraction of key blocks in descending order of estimated importance.
We set $\rho=0.1$, corresponding to approximately 10\% of the key blocks.
Table~\ref{tab:query_key_compression_topk} reports attention recall under clustered and original token orderings, with full-query and full-key estimation serving as the reference.

\begin{table}[H]
    \centering
    \caption{Top-k block-retrieval attention recall at a fixed retrieval ratio ($\rho=0.1$) under different query/key representations. Mean, P05, and Worst denote the mean, fifth percentile, and minimum attention recall across all attention heads and layers, respectively. All values are percentages. Higher is better.}
    \label{tab:query_key_compression_topk}
    \small
    \setlength{\tabcolsep}{4.0pt}
    \renewcommand{\arraystretch}{1.08}
    \begin{tabularx}{\linewidth}{@{}lll*{6}{>{\centering\arraybackslash}X}@{}}
        \toprule
        \multirow{2}{*}{Ordering} & \multirow{2}{*}{Query} & \multirow{2}{*}{Key}
        & \multicolumn{3}{c}{HunyuanVideo} & \multicolumn{3}{c}{Wan 2.1-14B} \\
        \cmidrule(lr){4-6}\cmidrule(l){7-9}
        & & & Mean & P05 & Worst & Mean & P05 & Worst \\
        \midrule
        \rowcolor{denserow}
        \cellcolor{white} & Full & Full & 77.05 & 27.41 & 18.97 & 73.48 & 32.81 & 19.66 \\
        & Centroid & Centroid & 75.37 & 27.27 & 18.96 & 72.35 & 32.78 & 19.66 \\
        & Centroid & Full & 76.14 & 27.38 & 18.96 & 72.88 & 32.80 & 19.66 \\
        \rowcolor{parkrow}
        \cellcolor{white}\multirow{-4}{*}{Clustered}
        & Full & Centroid & \textbf{76.50} & 27.29 & 18.96 & \textbf{73.12} & 32.79 & 19.66 \\
        \midrule
        \rowcolor{denserow}
        \cellcolor{white} & Full & Full & 61.90 & 18.93 & 12.73 & 56.49 & 21.17 & 10.51 \\
        & Centroid & Centroid & 56.78 & 17.56 & 08.46 & 54.78 & 20.52 & 10.51 \\
        & Centroid & Full & 56.91 & 18.91 & 12.72 & 53.17 & 21.13 & 10.51 \\
        \rowcolor{parkrow}
        \cellcolor{white}\multirow{-4}{*}{Original}
        & Full & Centroid & \textbf{56.94} & 17.49 & 08.59 & \textbf{54.84} & 20.66 & 10.51 \\
        \bottomrule
    \end{tabularx}
\end{table}

\section{Limitations}
This work focuses on accelerating video generation with diffusion models.
Our evaluation is limited to this setting and does not establish the applicability of PARK to text-based large language models or other multimodal tasks.
These settings may exhibit different attention patterns and inference workflows, which may require adaptations to PARK.
Further research is needed to determine whether PARK can provide similar efficiency benefits in these settings while preserving output quality.
\newpage

\section{Visualization of the Generated Videos}
\begin{figure}[H]
    \centering
    \includegraphics[width=\linewidth]{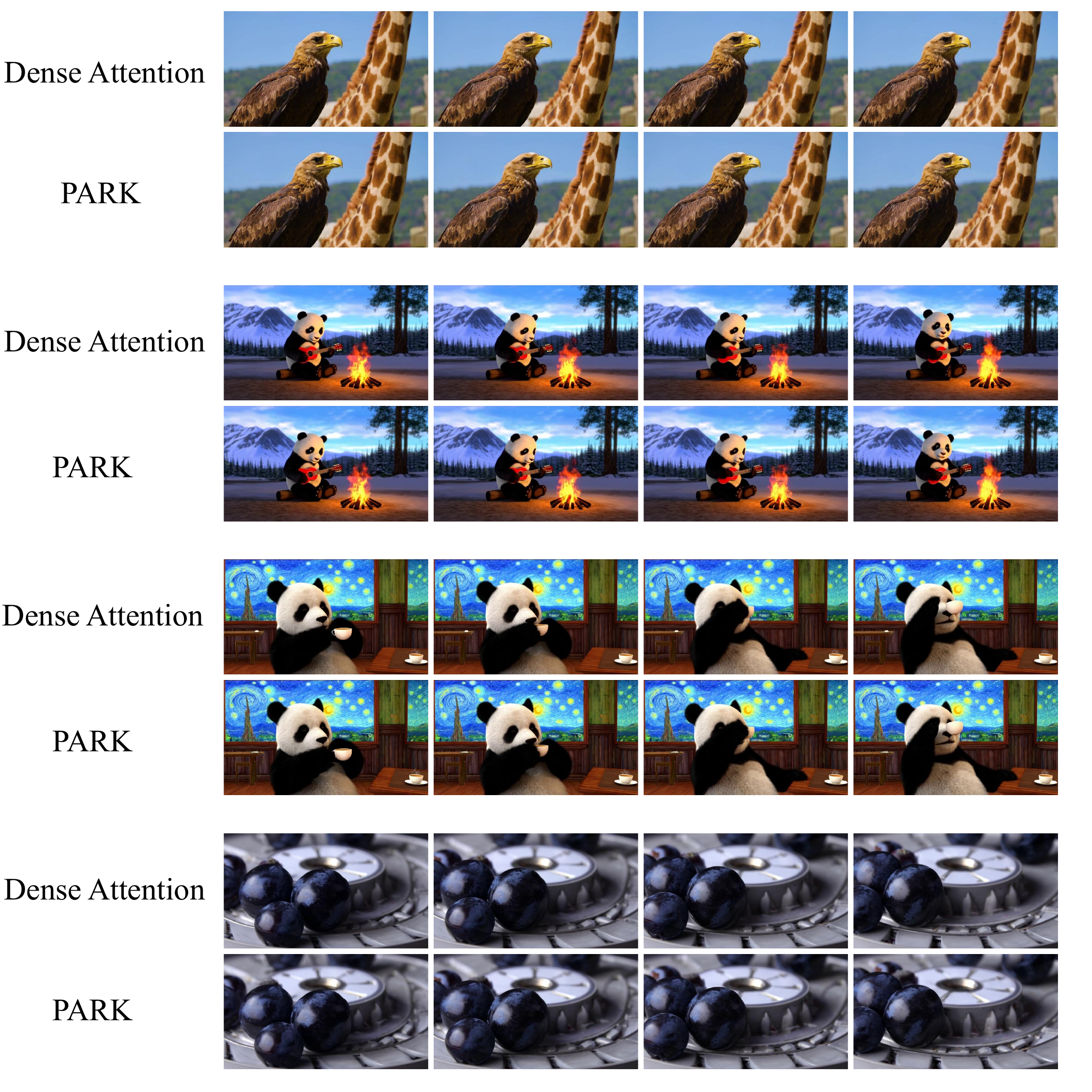}
    \caption{
        Comparison of Dense Attention and PARK for text-to-video generation with HunyuanVideo.
    }
    \label{fig:video_vis_hunyuanvideo}
\end{figure}

\begin{figure}[p]
    \centering
    \includegraphics[width=\linewidth]{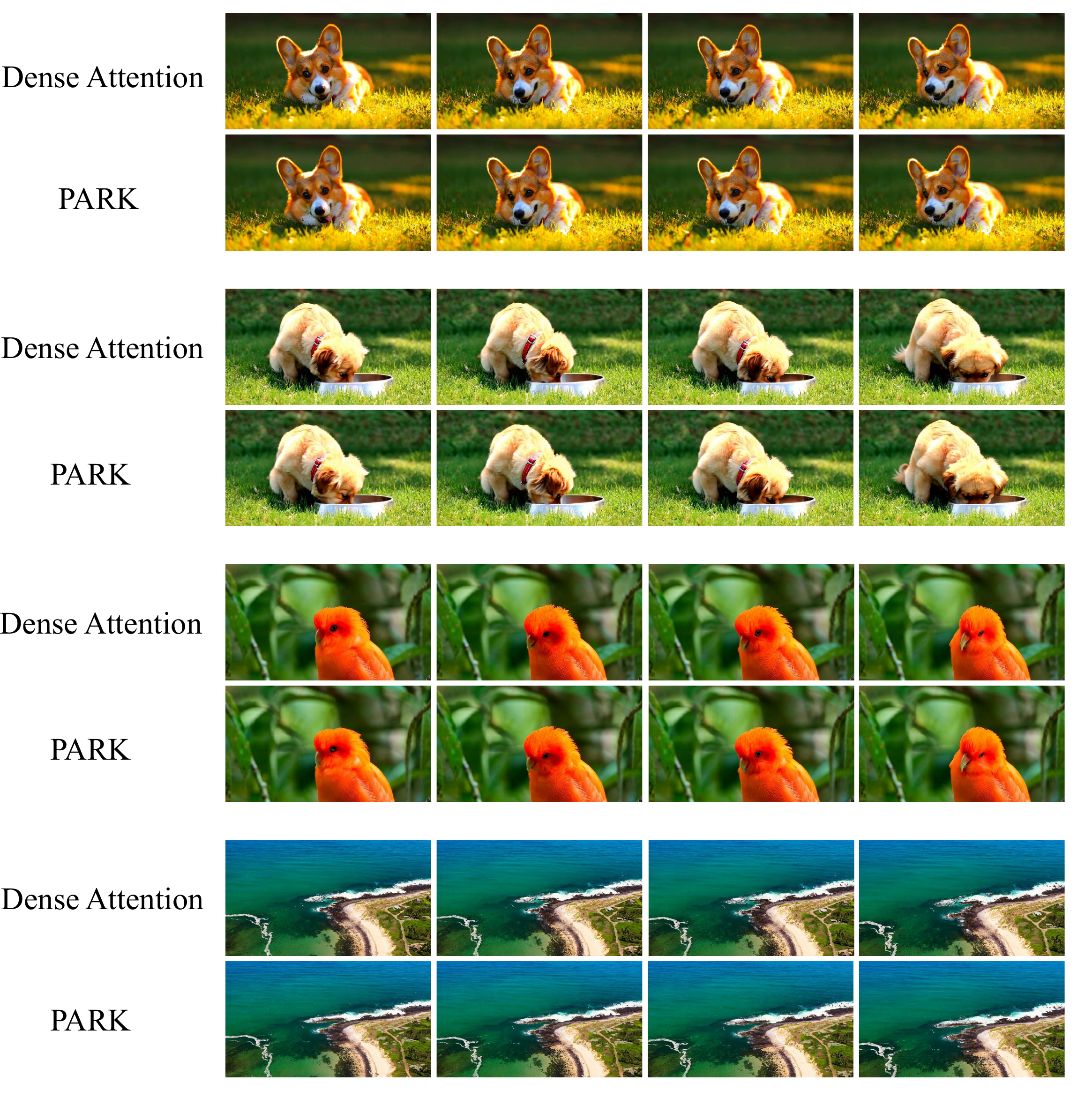}
    \caption{
        Comparison of Dense Attention and PARK for text-to-video generation with Wan2.1-14B.
    }
    \label{fig:video_vis_wan2.1}
\end{figure}

\begin{figure}[p]
    \centering
    \includegraphics[width=\linewidth]{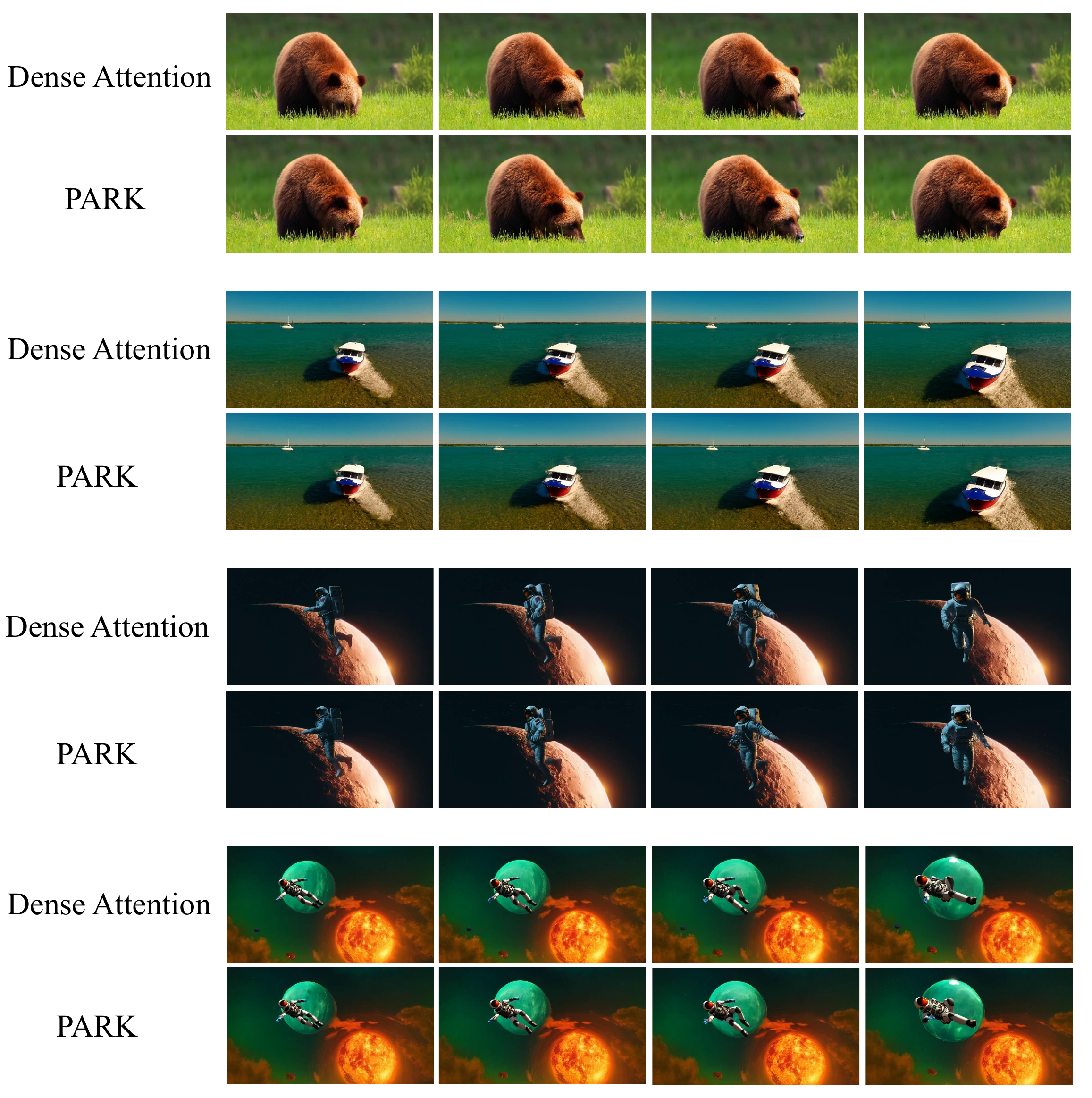}
    \caption{
        Comparison of Dense Attention and PARK for text-to-video generation with Wan2.1-1.3B.
    }
    \label{fig:video_vis_wan2.1-1.3B}
\end{figure}

\begin{figure}[p]
    \centering
    \includegraphics[width=\linewidth]{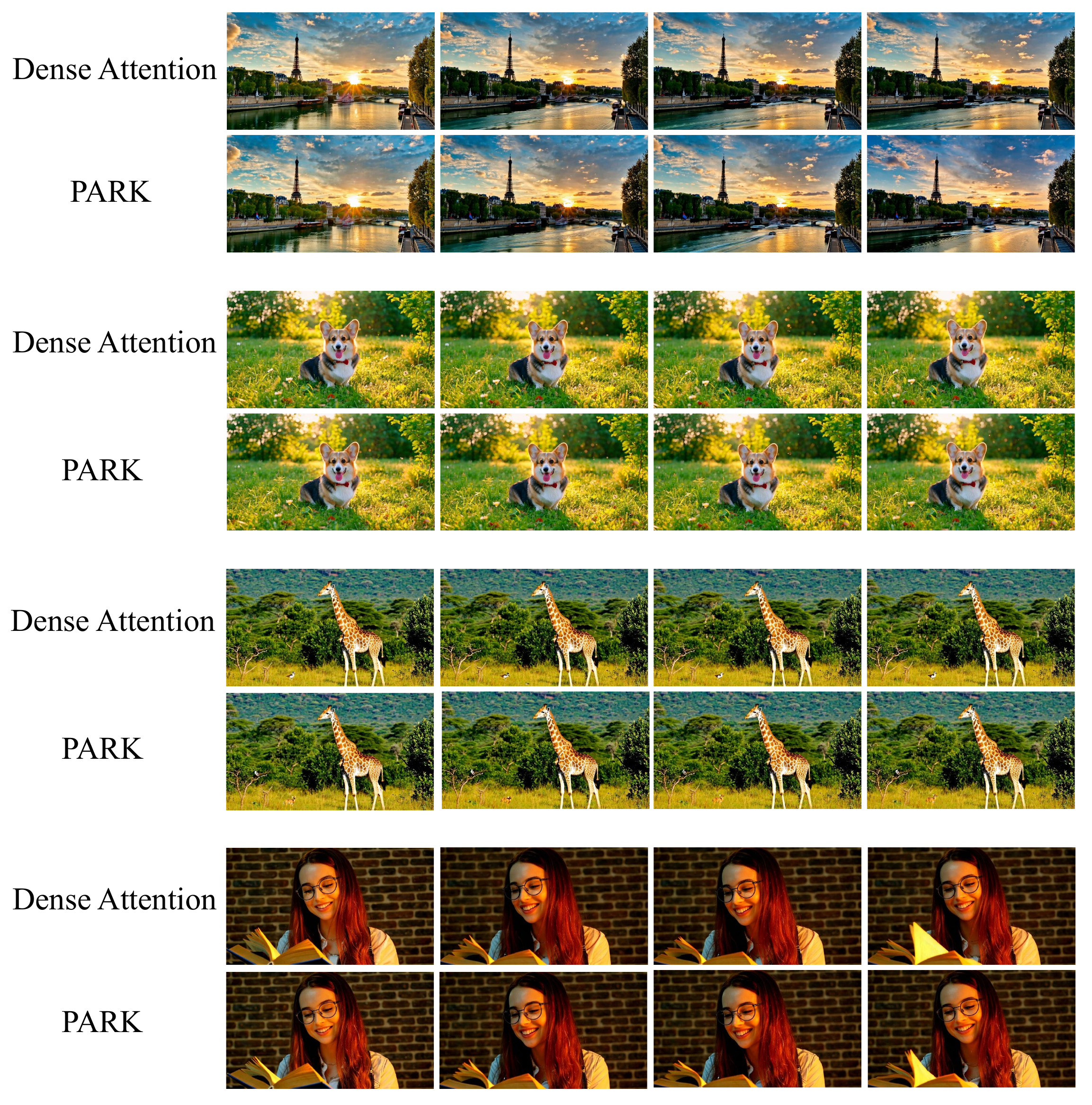}
    \caption{
        Comparison of Dense Attention and PARK for text-to-video generation with Wan2.2-14B.
    }
    \label{fig:video_vis_wan2.2}
\end{figure}

We provide visual comparisons between PARK and Dense Attention on HunyuanVideo, Wan2.1-14B, Wan2.1-1.3B, and Wan2.2-14B.
The results in Figures~\ref{fig:video_vis_hunyuanvideo}, \ref{fig:video_vis_wan2.1}, \ref{fig:video_vis_wan2.1-1.3B}, and~\ref{fig:video_vis_wan2.2} show that PARK preserves high pixel-level fidelity and achieves generation quality similar to that of dense attention.
More video samples are provided in the supplementary materials.
\vspace{30pt}

\end{document}